\documentclass[11pt]{article}

\usepackage[preprint]{acl}

\usepackage{times}
\usepackage{latexsym}
\usepackage{booktabs}
\usepackage{multirow}
\usepackage{makecell}
\usepackage[table]{xcolor}
\usepackage[T1]{fontenc}

\usepackage[utf8]{inputenc}

\usepackage{microtype}

\usepackage{inconsolata}

\usepackage{graphicx}
\usepackage{placeins}

\usepackage{amsthm}
\usepackage{times}
\usepackage{latexsym}
\usepackage[T1]{fontenc}
\usepackage[utf8]{inputenc}
\usepackage{microtype}
\usepackage{color}
\usepackage{amsmath}
\usepackage{amsmath,amssymb,amsfonts}
\usepackage{booktabs,multirow,array}
\usepackage{graphicx}
\usepackage{enumitem}
\usepackage{xcolor}
\usepackage{url}
\usepackage{tabularx}
\usepackage[most]{tcolorbox}
\usepackage{xcolor}
\usepackage{xspace}

\definecolor{AItitlegray}{RGB}{100,100,100}
\definecolor{AIgrayline}{RGB}{190,190,190}
\definecolor{BaseErrRed}{RGB}{180,50,50}
\definecolor{SteerFixGreen}{RGB}{30,120,76}
\definecolor{AIBg}{RGB}{252,252,252}
\definecolor{lightyellow}{RGB}{255, 249, 227}
\definecolor{lightgreen}{RGB}{235, 252, 235}
\definecolor{lightblue}{RGB}{230, 245, 255}
\newtcolorbox{AIBox}[1][]{%
  enhanced,
  colback=AIBg,
  colframe=AIgrayline,
  boxrule=0.5pt,
  arc=2pt,
  left=6pt, right=6pt, top=4pt, bottom=4pt,
  fonttitle=\bfseries\small,
  coltitle=white,
  colbacktitle=AItitlegray,
  title=#1,
}

\newcommand{\PromptSection}[1]{%
  \vspace{0.3em}\noindent\textbf{#1}\vspace{0.2em}%
}

\definecolor{ourcolor}{HTML}{8B4513}
\newcommand{\ours}{\textcolor{ourcolor}{\textsc{Lrs}}\xspace}

\title{Latent Reward Steering: An Adaptive Inference-Time Framework that Implicitly Promotes Cognitive Behaviors in Reasoning LLMs}

\newcommand{\equalcontrib}{\textsuperscript{*}}
\newcommand{\corresponding}{\textsuperscript{\ensuremath{\dagger}}}

\author{
\textbf{Jiakang Li}\equalcontrib\textsuperscript{1},
\textbf{Guanyu Zhu}\equalcontrib\textsuperscript{2},
\textbf{Can Jin}\equalcontrib\textsuperscript{1},
\textbf{Chenxi Huang}\textsuperscript{3},
\textbf{Dexu Yu}\textsuperscript{4},
\textbf{Ronghao Chen}\textsuperscript{5}
\\
\textbf{Yang Zhou}\textsuperscript{1},
\textbf{Hongwu Peng}\textsuperscript{6},
\textbf{Xuanqi Lan}\textsuperscript{7},
\textbf{Dimitris N. Metaxas}\corresponding\textsuperscript{1},
\textbf{Youhua Li}\corresponding\textsuperscript{8}
\\
\\
\textsuperscript{1}Rutgers University \quad
\textsuperscript{2}South China Agricultural University \quad
\textsuperscript{3}Columbia University
\\
\textsuperscript{4}Fenz.AI \quad
\textsuperscript{5}QuantaAlpha \quad
\textsuperscript{6}Adobe
\\
\textsuperscript{7}Santa Clara University \quad
\textsuperscript{8}City University of Hong Kong
\\
Contact: \texttt{\{jiakang.li@rutgers.edu\}}
\\
\small{\equalcontrib Equal contribution. \quad \corresponding Equal corresponding authors.}
}

\begin{document}
\maketitle
\begin{abstract}

Strong reasoning depends not only on model knowledge but also on how effectively cognitive behaviors are deployed during generation. Existing methods often rely on explicit behavior-level control, making them insufficiently adaptive when failures and required corrections vary across reasoning states, tasks, and models. To this end, we propose Latent Reward Steering (\ours), an adaptive inference-time framework that promotes cognitive behaviors by optimizing the sparse-autoencoder (SAE) latent states that implicitly carry them.  Rather than relying on predefined cognitive behaviors or steering directions derived from them, \ours trains a latent reward model on reasoning traces by final answer correctness to estimate the quality of intermediate latent states. During inference, reward gradients provide state-specific correction directions for fragile latent states, while a reward and confidence gate restricts intervention to states the reward signal flags as fragile. Experiments on multiple reasoning LLM backbones and benchmarks show that \ours consistently improves performance over various baselines, and post-hoc analyses further indicate that \ours implicitly promotes good cognitive behaviors that fix the original reasoning errors. Code is available at: \textcolor{blue}{\url{https://github.com/jiakanglee/Latent-Reward-Steering}}.
\end{abstract}

\section{Introduction}
\label{introduction}

Performing step-by-step reasoning to solve complex problems has become a central research focus in large language models \citep{wei2022chain, kojima2022large}. Yet even strong reasoning models remain brittle: a single early mistake, such as a flawed assumption or a skipped verification step, can gradually derail an otherwise promising reasoning chain~\citep{gan2025rethinking, huang2023large, tyen2024llms}. Recent work highlights cognitive behaviors such as verification, backtracking, and subgoal setting as important ingredients of successful reasoning \citep{gandhi2025cognitive}. This suggests that some reasoning failures are not purely failures of model knowledge, but failures to induce cognitive behaviors at the right moments within an ongoing reasoning chain.

\begin{figure}[t]
    \centering
    \includegraphics[width=0.98\linewidth]{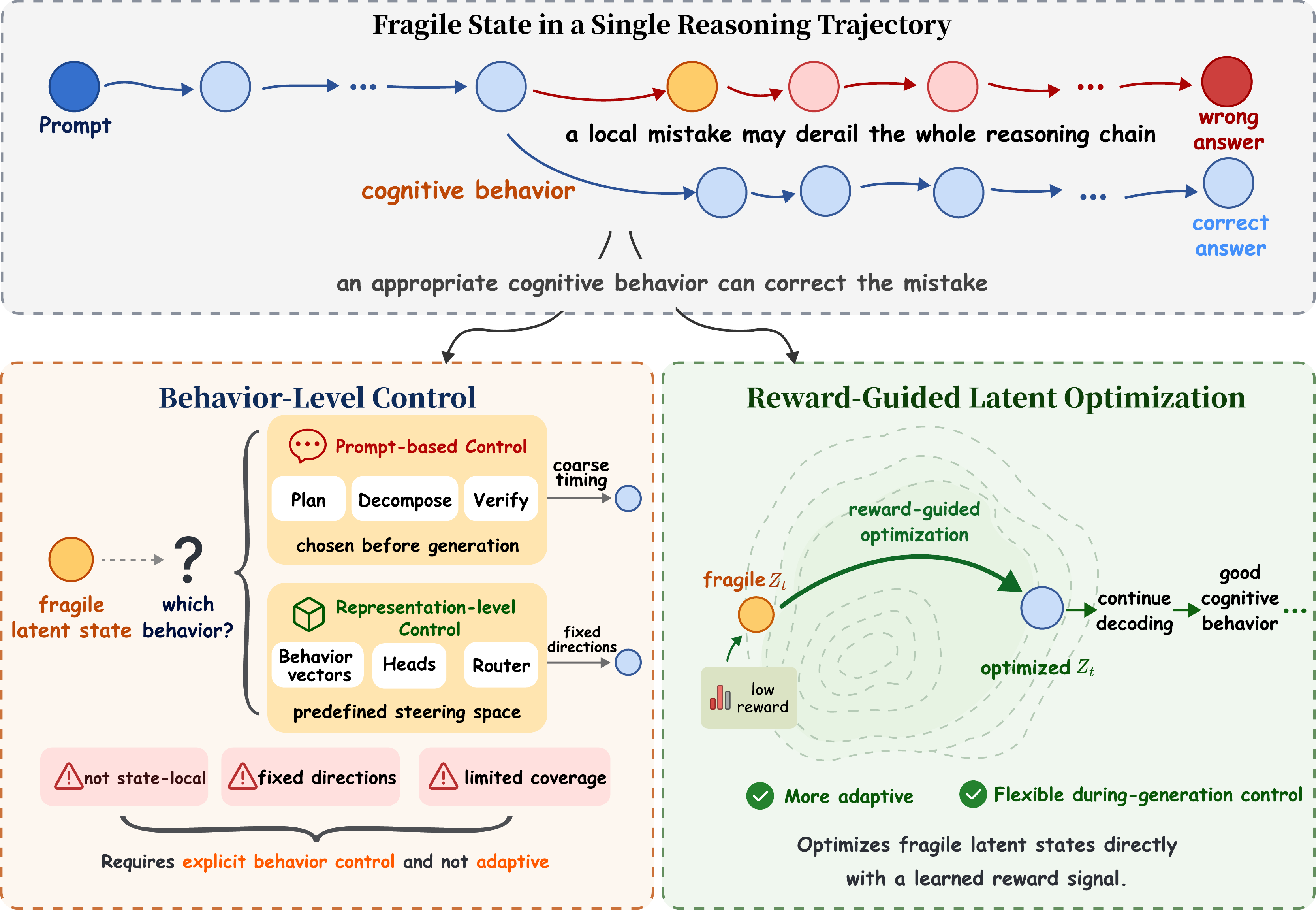}
\caption{Motivation. A fragile reasoning state can derail reasoning, while explicit behavior-level control may suffer from fixed labels and directions. \ours instead optimizes fragile latent states with a learned reward signal and implicitly promotes useful cognitive behaviors.}
    \label{fig:lrs_motivation}
\vspace{-2mm}
\end{figure}

The importance of cognitive behaviors in reasoning LLMs has motivated a line of work on controlling such behaviors, most of which involve explicit behavior-level control. Prompt-based methods elicit desired cognitive behaviors through textual instructions, from few-shot in-context learning \citep{brown2020} and chain-of-thought (COT) prompting \citep{wei2022chain, kojima2022large} to more specific behaviors such as sub-goal decomposition \citep{zhou2022least, wang2023plan}, strategic planning \citep{ zheng2024take}, and verification \citep{weng2023large, miao2023selfcheck, dhuliawala2024chain}. Representation-level steering methods instead intervene directly on latent states \citep{turner2023steering, zou2023representation}, by associating cognitive behaviors with steering directions \citep{chen2025seal}, head-specific interventions \citep{zhang2025understanding}, or routed behavior-vector libraries \citep{ye2026riser}.

Compared to prompt-based methods, which explicitly designate particular cognitive behaviors in text before generation begins and cannot target the step where an error actually occurs, representation-level steering methods intervene directly during decoding, offering a more direct and promising space for cognitive behavior control. However, this advantage in interface does not translate into adaptivity: representation-level methods still follow the same explicit behavior-level paradigm as prompt-based ones, where the behaviors to control are predefined and represented as behavior-specific intervention objects such as steering directions, selected heads, or vector libraries. Such a paradigm is not adaptive, since it relies on predefined cognitive behaviors (e.g., verification, backtracking) that may not apply uniformly across models \citep{gandhi2025cognitive}, and on steering directions derived from these predefined behaviors that may not match the local reasoning state \citep{chen2025seal}. As a result, although latent-level intervention is a promising direction, both prompt-based and representation-level methods remain insufficiently adaptive when failures and required corrections vary across reasoning tasks and models. 

This motivates an interesting research question: \emph{Can we promote good cognitive behaviors adaptively at the latent level without committing to predefined behaviors or their derived steering directions?} Recent findings make this question plausible. First, sparse latent states can be viewed as an internal space where the model's ongoing deployment of cognitive behaviors is implicitly represented \citep{wang2026beyond}. Second, useful cognitive mechanisms often already exist in the model's latent space, and gains can come from deploying them better rather than from injecting new behaviors \citep{venhoff2025base, ward2025reasoning}. Third, recent work shows that latent representations themselves encode reward-like quality signals that can be recovered by a learned model \citep{du2025latent}. Building on these observations, we therefore hypothesize that reward-guided optimization of latent states can adaptively promote the good cognitive behaviors already encoded in latent states during reasoning, without ever using explicit behavior control. 

Based on this hypothesis, we propose \ours, an adaptive inference-time framework that promotes cognitive behaviors by directly optimizing SAE latent states \citep{cunningham2023sparse, templeton2024scaling} with a learned reward signal. Rather than relying on predefined cognitive behaviors or steering directions derived from them, \ours trains a latent reward model on successful and unsuccessful reasoning traces to estimate the quality of intermediate latent states. During inference, the reward gradient supplies a state-specific correction direction for the current latent, while a reward and confidence gate restricts intervention to states the reward signal flags as fragile-prone, helping preserve reasoning steps that are already likely to be healthy. 

Our main contributions are summarized below:
\begin{itemize}
    \setlength{\itemsep}{0pt}
    \setlength{\parskip}{0pt}
    \setlength{\parsep}{0pt}
    \setlength{\topsep}{2pt}


 \item To the best of our knowledge, we are the first to frame cognitive behavior control for LLM reasoning as \emph{implicit latent-state optimization}, shifting the focus from explicitly selecting predefined behaviors to adaptively optimizing latent states that represent ongoing cognitive behavior deployment.

\item We introduce \textbf{\ours}, an adaptive inference-time framework that steers fragile SAE latent states through reward-guided correction together with reward and confidence gating, without relying on predefined cognitive behaviors or their derived steering directions.

\item We show that \ours consistently improves inference-time reasoning across multiple LLMs and challenging benchmarks, while qualitative and case-level analyses suggest that it implicitly promotes helpful cognitive behaviors such as solution verification and course correction.


\end{itemize}

\section{Related Work}

\paragraph{Inference-time reasoning and prompt-based behavior control.}
Inference-time reasoning has become an important way to improve LLM performance on complex tasks. Few-shot in-context learning and COT prompting elicit general step-by-step reasoning behavior \citep{brown2020, wei2022chain, kojima2022large,jin2025apeer}. Later prompting methods target more specific cognitive behaviors, including sub-goal decomposition \citep{zhou2022least, wang2023plan,jin2025two}, strategic planning \citep{zheng2024take}, and verification \citep{weng2023large, miao2023selfcheck, dhuliawala2024chain,jin2025your,zhang2026cm2}. 

\paragraph{Representation-level steering for reasoning.}
Activation steering and representation engineering provide a more direct way to influence model behavior by modifying internal states during generation \citep{zhang2026locate,turner2023steering, zou2023representation,jin2026reasoning,zhang2026soft}. Recent work applies this idea to reasoning control. SEAL decomposes reasoning traces into components such as execution, reflection, and transition, and learns steering vectors to calibrate them \citep{chen2025seal}. CREST identifies attention heads associated with behaviors such as verification and backtracking, and derives head-specific steering directions \citep{zhang2025understanding}. RISER builds a reusable library of reasoning vectors and learns a router to compose them during inference \citep{ye2026riser}. These methods make representation-level intervention a promising interface for cognitive behavior control, but remain tied to predefined behaviors, selected heads, fixed directions, or finite vector libraries.

\paragraph{Cognitive behaviors and implicit latent-state optimization.}
Recent studies highlight the role of cognitive behaviors in LLM reasoning. Cognitive behaviors such as verification, backtracking, and subgoal setting are important ingredients of strong reasoning performance \citep{gandhi2025cognitive}. Other work suggests that useful reasoning mechanisms may already exist in base models, and that gains can come from better deployment of these mechanisms rather than from adding new knowledge \citep{venhoff2025base, ward2025reasoning, venhoff2025steering}. These findings motivate cognitive behavior control, but also reveal the limitation of explicit behavior-level methods: predefined behaviors may not apply uniformly across models, and fixed intervention directions may not match the current reasoning state. In contrast, \ours frames cognitive behavior control as implicit latent-state optimization, using latent states to adaptively steer fragile reasoning states during decoding.

\section{Method}

\subsection{Problem Setup}
 \label{sec:problem-setup}

\begin{figure*}[t]
    \centering
    \includegraphics[width=0.95\textwidth]{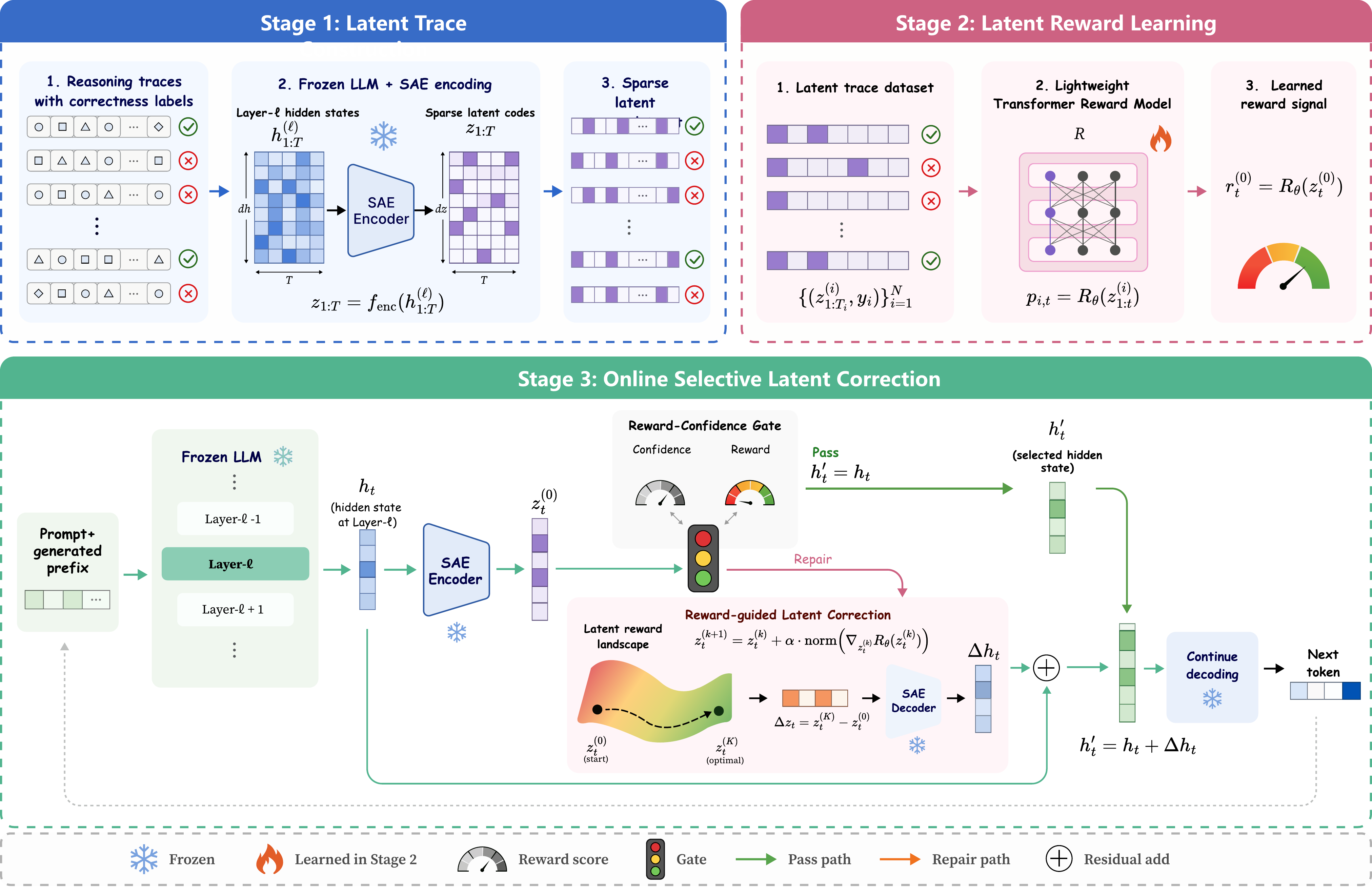}
    \caption{Framework of \ours. It first constructs SAE latent traces from reasoning trajectories, then trains a latent reward model to estimate intermediate-state quality. During inference, a reward and confidence gate identifies fragile states, and reward-guided latent correction updates the hidden state before decoding continues.}
    \label{fig:method}
    \vspace{-3mm}
\end{figure*}



We formalize the problem of \emph{adaptive cognitive behavior promotion for LLM reasoning} as an inference-time intervention on the model's latent states. Let a parameter-frozen reasoning model $\mathcal{M}$ process an input prompt $x$ and autoregressively generate a response $y = (y_1, \dots, y_T)$ token by token. At generation step $t$, let $h_t \in \mathbb{R}^{d_h}$ denote the hidden activation at layer $\ell$. As motivated in Section~\ref{introduction}, we view $h_t$ as an internal state that implicitly carries the model's intrinsic deployment of cognitive behaviors.

\paragraph{SAE-based intervention space.}
Directly steering the dense activation $h_t$ is difficult because it is high-dimensional and entangled. We therefore use a pretrained sparse autoencoder (SAE) to map $h_t$ into a low-dimensional sparse latent representation:
\begin{equation}
    z_t = f_{\mathrm{enc}}(h_t), \qquad \hat{h}_t = f_{\mathrm{dec}}(z_t),
\end{equation}
where $z_t \in \mathbb{R}^{d_z}$ is a $d_z$-dimensional sparse code. We use the pretrained SAEs released by \citet{venhoff2025base}, which were trained on each reasoning model's hidden activations and provide a model-specific sparse code: $d_z=10$ for Open-Reasoner-7B \cite{hu2026open} and $d_z=5$ for Open-Reasoner-1.5B \cite{hu2026open}. Prior work suggests that SAE latents expose interpretable internal features in language models \citep{cunningham2023sparse, templeton2024scaling}, and that SAE latent dimensions can be used to identify and analyze reasoning-relevant cognitive behaviors such as verification, backtracking, and constraint checking in the model's internal states \citep{venhoff2025base, ward2025reasoning, wang2026beyond}. We therefore use SAE latents as the intervention space for \ours.

\paragraph{Inference-time steering objective.}
Given a decoding step $t$, our goal is to obtain a corrected latent state $z'_t = z_t + \Delta z_t$ that improves the likelihood of a correct final answer when decoding continues from the corresponding hidden state. The update is adaptive to the current state, free from predefined cognitive behavior labels or behavior-specific steering directions, and selective enough to avoid disrupting already healthy reasoning states.


\subsection{Method Overview}

As shown in Figure~\ref{fig:method}, \ours has three stages. It first constructs SAE latent reasoning traces from a frozen reasoning model and labels each trace by final-answer correctness. It then trains a latent reward model to estimate the quality of intermediate latent states from successful and unsuccessful traces, without using explicit cognitive behavior annotations. During inference, \ours uses the learned reward signal to identify fragile states and applies reward-guided latent correction only when the reward and confidence gate triggers intervention. The decoded latent residual is added back to the hidden activation, enabling adaptive steering of fragile reasoning states without predefined behavior labels or behavior-specific directions.

\subsection{Latent Reward Training}
We first construct a dataset of sparse latent reasoning traces. For each solved example, we store a latent sequence together with a binary label indicating whether the final answer is correct. Formally, each example is represented as
\begin{equation}
    \mathcal{S}_i = (z_{i,s_i}, z_{i,s_i+1}, \dots, z_{i,T_i}), \qquad y_i \in \{0,1\},
\end{equation}
where $s_i$ is the beginning of the reasoning segment.

We train a lightweight Transformer reward model $R_\theta$ over these latent sequences:
\begin{equation}
    p_{i,t} = R_\theta(z_{i,s_i:t}) \in (0,1),
\end{equation}
where $p_{i,t}$ is the predicted probability that the reasoning trace belongs to a correct sample. In implementation, the binary label $y_i$ is repeated over all positions of the sequence, and the model is trained using binary cross-entropy:
\begin{equation}
    \mathcal{L}_{\mathrm{RM}} = -\sum_i \sum_{t=s_i}^{T_i} \Big[y_i \log p_{i,t} + (1-y_i)\log(1-p_{i,t})\Big].
\end{equation}
To mitigate class imbalance, we apply weighted random sampling during training.

The reward model itself is a small Transformer encoder with LayerNorm on the latent input, a learned embedding layer, positional encoding, two Transformer blocks, and an MLP head. 

\subsection{Online Reward-Guided Latent Correction}
\label{sec:steering}
At generation time, we intervene only on generation tokens, not on the prompt prefix. Given the current activation $h_t$, we encode it into a sparse latent vector built from the SAE encoder:
\begin{equation}
    z_t^{(0)} = f_{\mathrm{enc}}(h_t).
\end{equation}
We then evaluate the reward model on the current latent (implemented as a sequence of length $1$ for efficiency) and obtain an initial reward score
\begin{equation}
    r_t^{(0)} = R_\theta(z_t^{(0)}).
\end{equation}
If steering is triggered at this step ($\mathbb{I}_{\mathrm{steer}}(t) = 1$, with the gating rule defined in Section~\ref{sec:gate}), we optimize the latent by normalized gradient ascent for $K$ steps:
\begin{equation}
    z_t^{(k+1)} = z_t^{(k)} + \alpha \frac{\nabla_{z_t^{(k)}} R_\theta(z_t^{(k)})}{\left\|\nabla_{z_t^{(k)}} R_\theta(z_t^{(k)})\right\|_2 + \varepsilon},
\end{equation}
where $\alpha$ is the step size and $K$ is the number of steering iterations.

Instead of decoding the entire optimized latent, we only decode the latent \emph{difference}
\begin{equation}
    \Delta z_t = z_t^{(K)} - z_t^{(0)},
\end{equation}
and project it back to activation space with the SAE decoder matrix $W_{\mathrm{dec}}$:
\begin{equation}
    \Delta h_t = \Delta z_t W_{\mathrm{dec}}, \qquad h_t' = h_t + \Delta h_t.
\end{equation}
The steered hidden state $h_t'$ is then fed into the subsequent layers of the LLM.

\subsection{Selective Reward and Confidence Gating}
\label{sec:gate}

Applying latent updates at every generation step may disrupt already-correct reasoning, as the ungated variant \textsc{\ours Basic} degrades on several benchmarks (Table~\ref{tab:lrs_results}). We therefore introduce a selective gate based on two signals: the current reward score $r_t = R_\theta(z_t^{(0)})$ and the previous-token decoding confidence $c_{t-1}$, defined as the maximum softmax probability at step $t-1$.

Steering is triggered when either signal suggests a fragile local state:
\begin{equation}
    \mathbb{I}_{\mathrm{steer}}(t) =
    \begin{cases}
        1, & \text{if } r_t < \tau_r, \\
        1, & \text{if } r_t \ge \tau_r \text{ and } c_{t-1} < \tau_c, \\
        0, & \text{otherwise},
    \end{cases}
\end{equation}
where $\tau_r$ and $\tau_c$ are the reward and confidence thresholds. Low reward indicates that the current latent state is unlikely to lead to a correct answer, while low decoding confidence suggests local uncertainty even when the reward is acceptable. Thus, \ours intervenes only when correction is needed and preserves states that appear already on track.



\section{Experiments}

In this paper, we study the following research questions:

\begin{itemize}[leftmargin=1.3em,itemsep=1pt,topsep=2pt,parsep=0pt,partopsep=0pt]
    \item \textbf{RQ1:} Can \ours improve reasoning performance?
    \item \textbf{RQ2:} Does the latent reward model provide a meaningful correction signal?
    \item \textbf{RQ3:} Do \ours-steered traces implicitly promote useful cognitive behaviors?
\end{itemize}

We further conduct ablation and sensitivity analysis to examine the role of selective intervention and steering strength.

\begin{table*}[t]
\centering
\renewcommand{\arraystretch}{1.18}
\setlength{\tabcolsep}{4.2pt}
\resizebox{\textwidth}{!}{
\begin{tabular}{@{}lccccc|ccccc@{}}
\toprule
\rowcolor{gray!10}
\multirow{2}{*}{\textbf{Dataset}}
& \multicolumn{5}{c|}{\textbf{\textsc{Open-Reasoner-7B}}}
& \multicolumn{5}{c}{\textbf{\textsc{Open-Reasoner-1.5B}}} \\
\cmidrule(lr){2-6} \cmidrule(lr){7-11}
\rowcolor{gray!10}
& \makecell{\textbf{Base}\\\textbf{0-shot}}
& \makecell{\textbf{CoT}\\}
& \makecell{\textbf{5-shot}\\}
& \makecell{\textbf{\textsc{\ours Basic}}\\\textbf{0-shot}}
& \makecell{\textbf{\textsc{\ours}}\\\textbf{0-shot}}
& \makecell{\textbf{Base}\\\textbf{0-shot}}
& \makecell{\textbf{CoT}\\}
& \makecell{\textbf{5-shot}\\}
& \makecell{\textbf{\textsc{\ours Basic}}\\\textbf{0-shot}}
& \makecell{\textbf{\textsc{\ours}}\\\textbf{0-shot}}\\
\midrule
\textsc{MATH-500}
& 79.4
& 81.4
& 81.0
& 83.0 {\scriptsize\textcolor{gray}{(+3.6)}}
& \textbf{83.8 {\scriptsize\textcolor{gray}{(+4.4)}}}
& 59.2
& 58.6
& 57.2
& 59.0 {\scriptsize\textcolor{gray}{($-$0.2)}}
& \textbf{60.8 {\scriptsize\textcolor{gray}{(+1.6)}}} \\
\textsc{AIME24}
& 16.7
& 13.3
& 13.3
& 16.7 {\scriptsize\textcolor{gray}{(+0.0)}}
& \textbf{26.7 {\scriptsize\textcolor{gray}{(+10.0)}}}
& 3.3
& 6.7
& 6.7
& 0.0 {\scriptsize\textcolor{gray}{($-$3.3)}}
& \textbf{13.3 {\scriptsize\textcolor{gray}{(+10.0)}}} \\
\textsc{AIME25}
& 16.7
& 13.3
& 10.0
& 20.0 {\scriptsize\textcolor{gray}{(+3.3)}}
& \textbf{26.7 {\scriptsize\textcolor{gray}{(+10.0)}}}
& 3.3
& 0.0
& 3.3
& 0.0 {\scriptsize\textcolor{gray}{($-$3.3)}}
& \textbf{6.7 {\scriptsize\textcolor{gray}{(+3.3)}}} \\
\textsc{GPQA-Diamond}
& 32.3
& 35.9
& 38.4
& 30.8 {\scriptsize\textcolor{gray}{($-$1.5)}}
& \textbf{39.4 {\scriptsize\textcolor{gray}{(+7.1)}}}
& 18.2
& 17.2
& 17.2
& 15.7 {\scriptsize\textcolor{gray}{($-$2.5)}}
& \textbf{22.7 {\scriptsize\textcolor{gray}{(+4.5)}}} \\
\textsc{AMC23}
& 50.0
& 55.0
& \textbf{65.0}
& 45.0 {\scriptsize\textcolor{gray}{($-$5.0)}}
& 60.0 {\scriptsize\textcolor{gray}{(+10.0)}}
& 30.0
& 32.5
& 30.0
& 32.5 {\scriptsize\textcolor{gray}{(+2.5)}}
& \textbf{37.5 {\scriptsize\textcolor{gray}{(+7.5)}}} \\
\textsc{IneqMath}
& 46.0
& 48.0
& 48.0
& 52.0 {\scriptsize\textcolor{gray}{(+6.0)}}
& \textbf{60.0 {\scriptsize\textcolor{gray}{(+14.0)}}}
& 29.0
& 30.0
& 28.0
& 28.0 {\scriptsize\textcolor{gray}{($-$1.0)}}
& \textbf{34.0 {\scriptsize\textcolor{gray}{(+5.0)}}} \\
\bottomrule
\end{tabular}}
\caption{Main results are all reported under greedy decoding with maximum token budget 4000. Values in parentheses denote absolute gains of \ours BASIC / \ours over the corresponding Base model. Accuracies and gains are rounded independently to one decimal place, with gains computed before rounding. CoT and few-shot columns report accuracies under chain-of-thought and few-shot prompting, respectively. \ours BASIC is ungated, while full \ours uses reward and confidence gating.}
\label{tab:lrs_results}
\vspace{-0.5em}
\end{table*}
\subsection{Experimental Setup}
\label{sec:experimental_setup}
We evaluate \ours with \textsc{Open-Reasoner-7B} as the primary base model and additionally include \textsc{Open-Reasoner-1.5B} to assess generalization across model scales. We adopt the pretrained SAE checkpoints released by \citet{venhoff2025base}: steering is applied at layer 20 in a 10-dimensional SAE latent space for \textsc{Open-Reasoner-7B}, and in a 5-dimensional SAE latent space for \textsc{Open-Reasoner-1.5B} \cite{hu2026open}. The main results in Table~\ref{tab:lrs_results} are reported for both backbones, while most qualitative and diagnostic analyses (reward-score separation, post-hoc cognitive behavior annotation, SAE interpretability, efficiency, and case studies) are conducted on the primary \textsc{Open-Reasoner-7B} model \cite{hu2026open}. 

We evaluate on six reasoning benchmarks: \textsc{MATH-500}~\citep{hendrycks2021measuring}, \textsc{AIME24}~\cite{aime2024_i,aime2024_ii}, \textsc{AIME25}~\cite{aime2025_i,aime2025_ii}, \textsc{AMC23}~\cite{amc23_hf}, \textsc{IneqMath}~\citep{sheng2026solving}, and \textsc{GPQA-Diamond}~\citep{rein2023gpqa}. All methods use greedy decoding and batch size $1$. Base, \ours Basic, and full \ours use zero-shot problem prompts and the CoT and five-shot baselines use the prompt templates described in Appendix~\ref{app:training_exp_details}.
More detailed training and experimental settings are provided in Appendix~\ref{app:training_exp_details}.

\subsection{Main Results: \ours Improves Reasoning Performance (RQ1)}
\label{sec:main_results}

\paragraph{\ours improves reasoning across model scales.}
As shown in Table~\ref{tab:lrs_results}, full \ours improves over standard zero-shot decoding on \emph{all} six reasoning benchmarks for \textsc{Open-Reasoner-7B}, with gains ranging from $+4.4$ on the \textsc{MATH-500} to $+14.0$ on the \textsc{IneqMath}, and we observe the same trend on the smaller same family model \textsc{Open-Reasoner-1.5B}, indicating that the method generalizes across model scales. These improvements all come from inference-time latent steering alone, without any training on the model weights.

\paragraph{Recovery exceeds degradation.}
To better understand \ours's effect on reasoning improvement at the trace level, we perform a matched-pair analysis between base and \ours reasoning chains ($N=398$, \textsc{Open-Reasoner-7B}). As shown in Table~\ref{tab:outcome_transition}, wrong-to-correct transitions account for 17.3\% of all matched examples, while correct-to-wrong transitions account for 7.8\%, yielding a net accuracy gain of 9.5 percentage points in this sample. Wrong-to-correct transitions are more than twice as frequent as correct-to-wrong transitions in the matched sample. These percentages use all matched examples as their denominator.

\paragraph{\ours consistently outperforms prompt-based behavior control.}
\ours outperforms both CoT and few-shot prompting on five of the six benchmarks for \textsc{Open-Reasoner-7B} and on all six for \textsc{Open-Reasoner-1.5B}. In contrast, prompt-based baselines are unstable: CoT and few-shot prompting underperform standard zero-shot decoding on several benchmarks (e.g., $-3.3$ for CoT on \textsc{AIME24} and \textsc{AIME25} with \textsc{Open-Reasoner-7B}), consistent with observations that reasoning-tuned models already internalize CoT-style reasoning behavior so adding explicit instructions does not necessarily help \citep{venhoff2025base}. These results empirically answer RQ1: \ours improves reasoning performance across models and benchmarks.

\begin{table}[t]
\centering
\renewcommand{\arraystretch}{1.1}
\setlength{\tabcolsep}{5pt}
\small
\begin{tabular}{lcc}
\toprule
\textbf{Transition} & \textbf{Base} $\rightarrow$ \textbf{\ours} & \textbf{Rate} \\
\midrule
Improved   & Wrong $\rightarrow$ Correct & 17.3\% \\
Degraded   & Correct $\rightarrow$ Wrong & 7.8\% \\
Preserved  & Correct $\rightarrow$ Correct & 27.4\% \\
Unresolved & Wrong $\rightarrow$ Wrong & 47.5\% \\
\bottomrule
\end{tabular}
\caption{Matched outcome transitions between base and \ours traces ($N=398$, \textsc{Open-Reasoner-7B}).}
\label{tab:outcome_transition}
\vspace{-3mm}
\end{table}

\begin{figure*}[t]
    \centering
    \includegraphics[width=\linewidth]{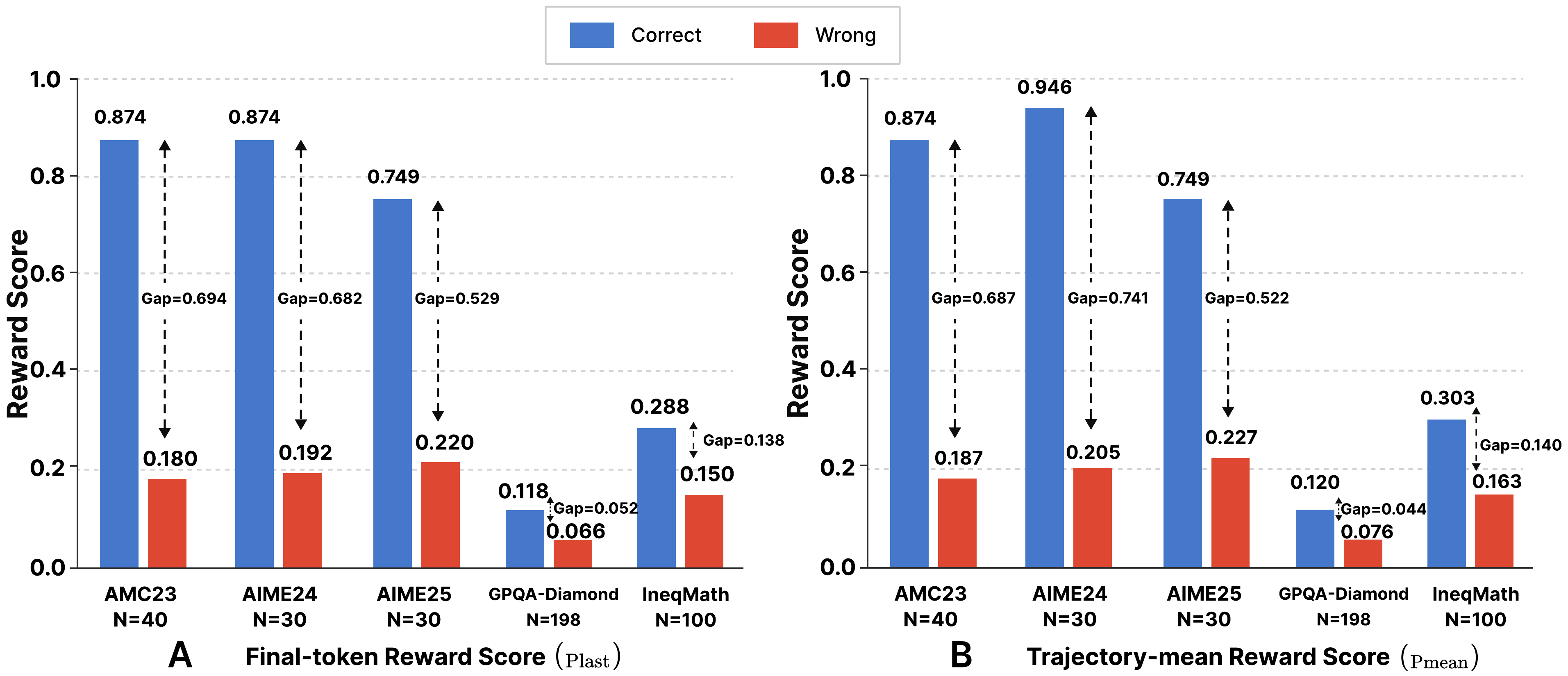}
    \caption{Latent reward scores distinguish successful and failed reasoning traces. We compare final-token reward $p_{\mathrm{last}}$ and trace-mean reward $p_{\mathrm{mean}}$ between correct and incorrect generations. 
    \vspace{-4mm}
    }
    \label{fig:reward_score_separation}
\end{figure*}

\subsection{Latent Reward Signal Quality (RQ2)}
\label{sec:reward_score_analysis}

We next examine whether the learned latent reward model captures a signal that is informative of reasoning quality.
For each generated reasoning trace, the reward model assigns token-level sigmoid scores over the SAE latent sequence.
Although trained only with trace-level correctness, the reward model uses token-level scores because local reasoning errors often propagate, making intermediate latent quality useful for selective correction.
We summarize these scores using two diagnostic statistics: the final-token reward $p_{\mathrm{last}}$, which reflects the reward estimate at the end of generation, and the trace-level mean reward $p_{\mathrm{mean}}$, which reflects the average quality signal across the reasoning process.

Figure~\ref{fig:reward_score_separation} shows that correct reasoning traces generally receive higher reward scores than incorrect reasoning traces across the evaluated datasets.
The separation is most pronounced on \textsc{AMC23}, \textsc{AIME24}, and \textsc{AIME25}, while \textsc{GPQA-Diamond} and \textsc{IneqMath} show the same ordering with smaller gaps.
This suggests that the reward model captures reasoning-quality differences in latent space.
Importantly, this analysis is not final-answer verification and not behavior classification: the reward model is trained only with final correctness labels over latent traces, without behavior annotations, answer-extraction labels, or predefined cognitive categories. This empirically answers RQ2, showing that the latent reward model provides a meaningful correction signal.

\subsection{\ours Implicitly Promotes Useful Cognitive Behaviors (RQ3)}
\label{sec:behavior_analysis}

We finally examine whether reward-guided latent steering is associated with interpretable changes in cognitive behavior.
This analysis is diagnostic only: behavior labels are never used to train the reward model, construct steering vectors, trigger the reward and confidence gate, or guide inference.
Instead, we annotate matched base and \ours reasoning chains post hoc to examine whether latent reward steering changes the observable reasoning process by promoting useful cognitive behaviors.
We conduct a matched-pair analysis between base and \ours reasoning chains ($N=398$, \textsc{Open-Reasoner-7B}) and annotate five cognitive behaviors following \citet{gandhi2025cognitive}: \emph{Strategic Planning}, \emph{Structured Decomposition}, \emph{Constraint Grounding}, \emph{Course Correction}, and \emph{Solution Verification}.
Annotations are produced by GPT-4o using a fixed post-hoc judging prompt, which is provided in Appendix~\ref{app:behavior_annotation}.
For each trace, a behavior is counted as present if it appears at least once.

\begin{figure}[htbp]
    \centering
    \includegraphics[width=0.9\linewidth]{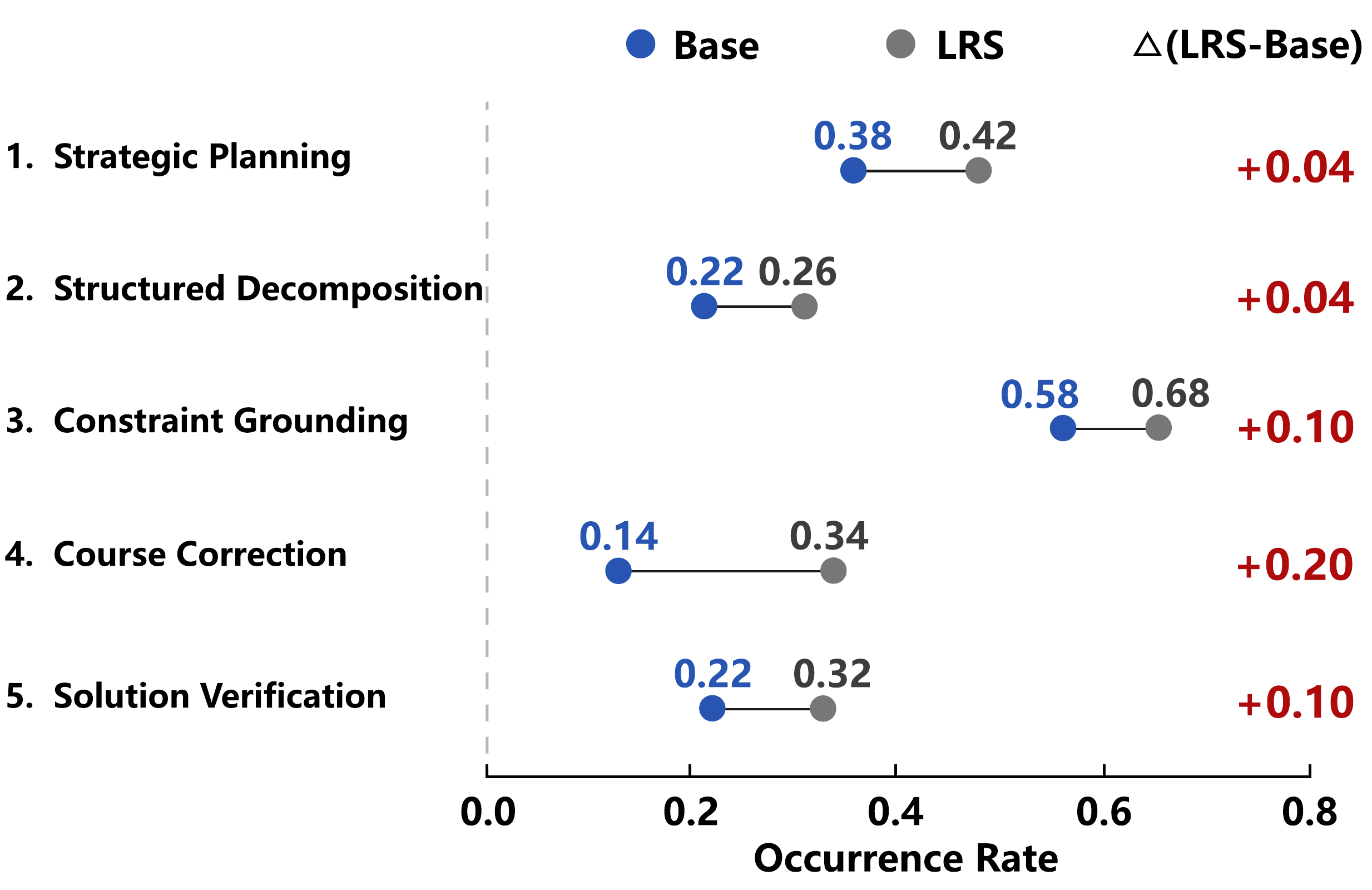}
    \caption{Post-hoc analysis shows that \ours-steered traces more often exhibit course correction, constraint grounding, and solution verification, with behavior labels used only for analysis.}
    \label{fig:cognitive_behavior_freq}
    \vspace{-2mm}
\end{figure}

\paragraph{Post-hoc behavior frequency.}
Figure~\ref{fig:cognitive_behavior_freq} reports the occurrence rates of the five annotated cognitive behaviors in matched base and \ours reasoning chains. \ours-steered traces show higher occurrence of \emph{Course Correction} ($+0.20$), \emph{Constraint Grounding} ($+0.10$), and \emph{Solution Verification} ($+0.10$). These behaviors are closely related to local reasoning correction: revisiting potentially wrong steps, checking consistency with problem constraints, and auditing the final derivation or answer. This pattern is consistent with the mechanism of \ours. Since \ours does not use behavior annotations, behavior-specific vectors, or predefined behavior triggers, these shifts should not be interpreted as explicit behavior selection. Rather, they suggest that reward-guided latent optimization changes fragile reasoning steps in ways associated with implicitly promoting more frequent helpful cognitive behaviors through latent-state optimization.

\paragraph{Case study.}
\begin{figure}[t]
\begin{AIBox}[Case Study: AIME25 Q2 --- Counting with Strict Constraints]
\footnotesize

\PromptSection{Question}
Nine players choose chocolate, vanilla, or strawberry, each flavor appears and the counts satisfy $c>v>s$. Count assignments modulo $1000$.

\PromptSection{Base Error}
The base sets up $c+v+s=9$ but keeps only $(5,3,1)$, giving $\binom{9}{5,3,1}=504$. It rejects $(6,2,1)$ and $(4,3,2)$ despite $6>2>1$ and $4>3>2$.

\PromptSection{\ours Correction}
\ours revisits the enumeration and keeps all valid partitions: $(6,2,1)$, $(5,3,1)$, and $(4,3,2)$, with counts $252$, $504$, and $1260$, hence $2016\equiv16\pmod{1000}$.

\PromptSection{Diagnostic Pathway}
Early reward and confidence interventions occur during count branching. After steering, the trace rechecks discarded cases and strict ordering.

\PromptSection{Interpretation}
The case illustrates \emph{Structured Decomposition}, \emph{Constraint Grounding}, and \emph{Course Correction}. SAE dimension names remain post-hoc priors, not deterministic causal labels.
\end{AIBox}
\caption{\ours repairs an incomplete enumeration by recovering the missed valid partitions.}
\label{fig:main_aime25_q2_case}
\vspace{-7mm}
\end{figure}

The compact case in Figure~\ref{fig:main_aime25_q2_case} illustrates the mechanism suggested by the aggregate analyses. The base reasoning commits to an incomplete enumeration and fails to revisit excluded cases, leading to an incorrect answer. In contrast, the \ours-steered trace receives early latent interventions during enumeration. After reward-guided optimization, SAE dimensions associated with algebraic execution and symbolic formatting become more active, and the trace later revisits the missing cases to produce a more complete solution. Additional qualitative cases with fuller diagnostics are provided in Appendix~\ref{app:case_studies} and Figure~\ref{fig:lrs_verification_case}. Together with the aggregate behavior-frequency analysis, these cases provide post-hoc evidence for RQ3: \ours-steered traces are associated with more frequent useful cognitive behaviors.

\vspace{-3mm}



\subsection{Additional Analyses: Stability, Interpretability, and Efficiency}
\label{sec:additional_analysis}

We include 3 supporting analyses to examine the stability, interpretability, and practical cost of \ours.

\begin{figure}[!bp]
    \centering
    \includegraphics[width=0.80\linewidth]{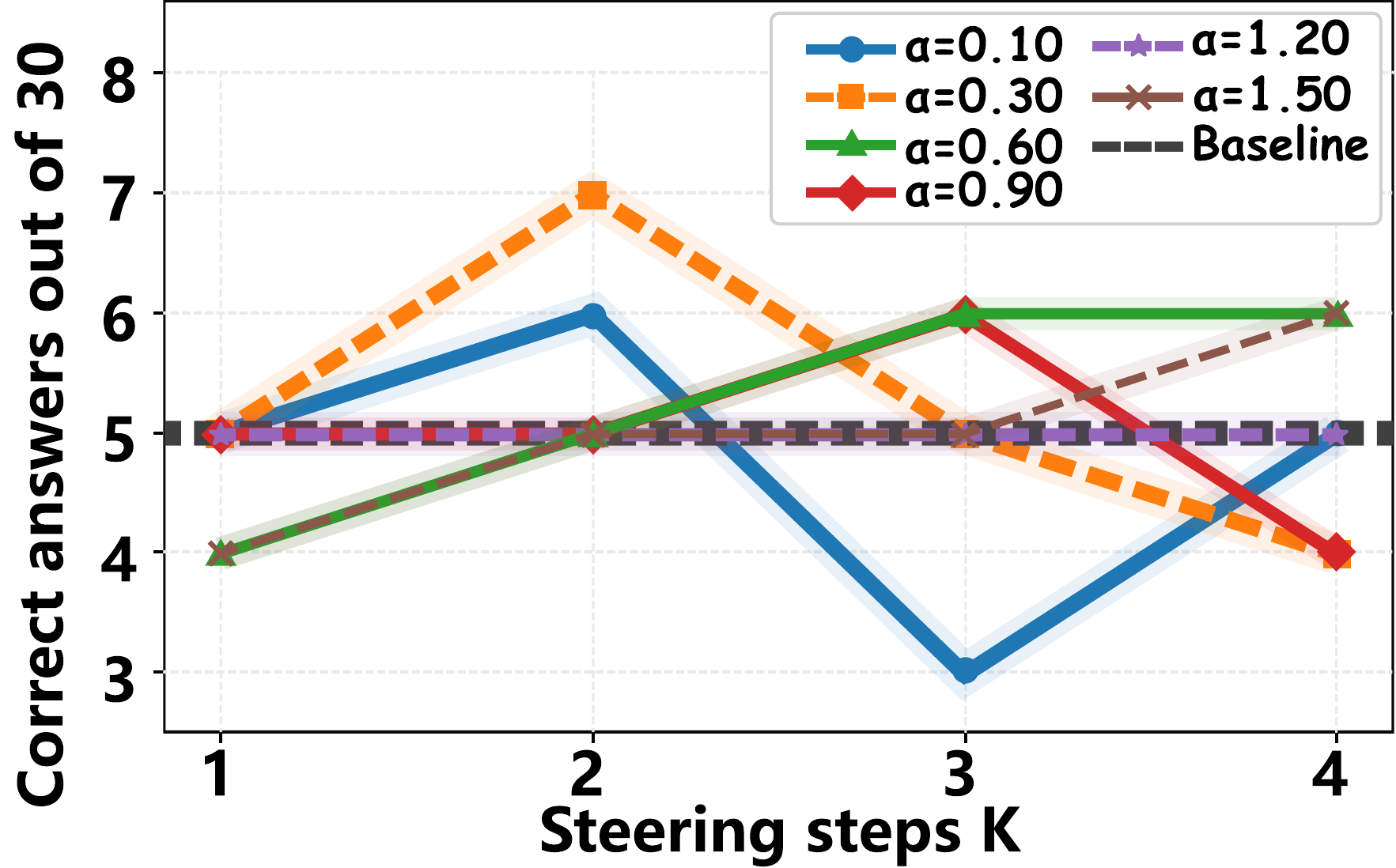}
    \caption{Steering-strength sensitivity on \textsc{AIME24}. Accuracy peaks under moderate updates rather than increasing monotonically and the dashed line indicates standard decoding.}
    \label{fig:steering_strength}
    \vspace{-2mm}
\end{figure}

\paragraph{Selective intervention and steering strength.}
Table~\ref{tab:lrs_results} shows that \textsc{\ours Basic} improves some datasets but degrades others, indicating that reward-guided gradients should not be applied indiscriminately. Full \textsc{\ours} uses the reward and confidence gate as a selective repair mechanism, and Figure~\ref{fig:steering_strength} shows that stronger updates do not monotonically improve accuracy and moderate intervention is more stable.

\paragraph{SAE interpretability.}
Table~\ref{tab:sae_interpretability} summarizes max-activating-context interpretations of the 10 SAE dimensions on \textsc{MATH-500}. These names provide post-hoc priors for analyzing latent changes, not deterministic mappings from dimensions to behaviors.

\begin{table}[!tbp]
\vspace{-1mm}
\centering
\renewcommand{\arraystretch}{1.05}
\setlength{\tabcolsep}{3pt}
\small
\resizebox{\columnwidth}{!}{
\begin{tabular}{ccl}
\toprule
\textbf{Dim.} & \textbf{Max Act.} & \textbf{Interpreted Pattern} \\
\midrule
$z_0$ & 0.218 & Geometric / structural modeling \\
$z_1$ & 0.287 & Theorem invocation / logical branching \\
$z_2$ & 0.477 & Algebraic flow / step execution \\
$z_3$ & 0.582 & Symbolic math / formatting \\
$z_4$ & 0.552 & Variable initialization / constant extraction \\
$z_5$ & 0.438 & Property definition / formula grounding \\
$z_6$ & 0.405 & Conclusion / answer consolidation \\
$z_7$ & 0.256 & Constraint checking / boundary validation \\
$z_8$ & 0.455 & Strategy selection / planning \\
$z_9$ & 0.283 & Complexity / asymptotic reasoning \\
\bottomrule
\end{tabular}
}
\caption{Compact interpretation of SAE latent dimensions from max-activating contexts on \textsc{MATH-500}.}
\label{tab:sae_interpretability}
\vspace{-2mm}
\end{table}

\paragraph{Inference efficiency.}
Table~\ref{tab:efficiency} reports the inference overhead of \ours. Average wall-clock time increases from 115.3s to 156.2s per problem ($1.35\times$), while the reward and confidence gate skips 72.1\% of tokens and steers 27.9\%.

\begin{table}[!tbp]
\centering
\vspace{-1mm}
\renewcommand{\arraystretch}{1.15}
\setlength{\tabcolsep}{5pt}
\small
\begin{tabular}{lcc}
\toprule
\textbf{Metric} & \textbf{Base} & \textbf{\ours} \\
\midrule
Avg.\ generated tokens                & 2595  & 2596 \\
Avg.\ wallclock / problem (s)         & 115.3  & 156.2 \\
Avg.\ generation cost (ms / token)    & 44.5  & 60.2 \\
Slowdown ratio                        & 1.00$\times$ & 1.35$\times$ \\
Steered tokens (\%)                   & ---  & 27.9 \\
Unsteered tokens (\%)                 & ---  & 72.1 \\
Avg.\ steering triggers / problem     & ---  & 725 \\
\quad Avg.\ triggers (correct answers) & --- & 526 \\
\quad Avg.\ triggers (wrong answers)   & --- & 924 \\
Avg.\ $\|\Delta z\|$ per steered token & ---  & 1.24 \\
SAE latent dimension                  & ---  & 10 \\
\bottomrule
\end{tabular}
\caption{Inference efficiency of \textsc{Base} and \textsc{\ours} on 200 problems from \textsc{AIME24}, \textsc{AIME25}, \textsc{AMC23}, and \textsc{IneqMath}.}
\label{tab:efficiency}
\vspace{-2mm}
\end{table}

\vspace{-2mm}

\section{Conclusion}
We presented \textbf{Latent Reward Steering}, an adaptive inference-time framework that improves LLM reasoning by applying reward-guided optimization to SAE latent states. \ours learns a latent reward model from successful and unsuccessful reasoning traces, uses reward gradients to correct fragile token-level states, and applies a reward--confidence gate to keep intervention selective. Experiments on \textsc{Open-Reasoner-7B} and \textsc{Open-Reasoner-1.5B} show consistent gains across reasoning benchmarks without updating model weights. Reward-score separation, post-hoc behavior analysis, and case studies further suggest that \ours-steered traces are associated with more frequent useful cognitive behaviors. These findings support latent-state optimization as a promising direction for inference-time reasoning improvement.

\vspace{20mm}

\section*{Limitations}
This work has several limitations as follows.
\begin{itemize}[leftmargin=1.3em, itemsep=1pt, topsep=2pt, parsep=0pt, partopsep=0pt]


     \item \textbf{Inference overhead.} \ours requires reward-model evaluation and latent-gradient updates during decoding. Selective gating reduces unnecessary interventions, but \ours remains slower than standard decoding.

    \item \textbf{Training--inference mismatch.} The reward model is trained on latent reasoning sequences but queried on local token-level states for efficiency. Future work could explore prefix-level or memory-augmented reward estimation.


\end{itemize}

\section*{Ethical considerations}

\ours performs inference-time latent steering based on a learned reward signal, which may introduce risks if the reward model is misaligned or poorly calibrated. In such cases, steering could amplify undesirable reasoning patterns, increase overconfidence in incorrect answers, or produce behavioral shifts that are difficult to interpret. Since latent-space interventions are less transparent than explicit prompting, their effects in open-ended and safety-sensitive scenarios remain uncertain. Our evaluation is limited to mathematical and scientific reasoning benchmarks, and does not fully characterize these risks. Future use of LRS should involve reward-model auditing, broader safety evaluation, and monitoring for unintended changes in model cognitive behavior.

\bibliography{custom}

@inproceedings{wei2022chain,
  title     = {Chain-of-Thought Prompting Elicits Reasoning in Large Language Models},
  author    = {Wei, Jason and Wang, Xuezhi and Schuurmans, Dale and Bosma, Maarten and Ichter, Brian and Xia, Fei and Chi, Ed and Le, Quoc and Zhou, Denny},
  booktitle = {Advances in Neural Information Processing Systems},
  year      = {2022}
}

@inproceedings{gandhi2025cognitive,
  title     = {Cognitive Behaviors that Enable Self-Improving Reasoners, or, Four Habits of Highly Effective STaRs},
  author    = {Gandhi, Kanishk and Chakravarthy, Ayush and Singh, Anikait and Lile, Nathan and Goodman, Noah D.},
  booktitle = {Conference on Language Modeling (COLM)},
  year      = {2025}
}

@inproceedings{chen2025seal,
  title     = {SEAL: Steerable Reasoning Calibration of Large Language Models for Free},
  author    = {Chen, Runjin and Zhang, Zhenyu and Hong, Junyuan and Kundu, Souvik and Wang, Zhangyang},
  booktitle = {Conference on Language Modeling (COLM)},
  year      = {2025}
}

@article{ye2026riser,
  title   = {RISER: Orchestrating Latent Reasoning Skills for Adaptive Activation Steering},
  author  = {Ye, Wencheng and Yuan, Xiaoyang and Bin, Yi and Jin, Hengyu and Peng, Liang and Zeng, Pengpeng and Shen, Heng Tao},
  journal = {arXiv preprint arXiv:2601.09269},
  year    = {2026}
}

@article{turner2023steering,
  title   = {Steering Language Models with Activation Engineering},
  author  = {Turner, Alexander M. and Thiergart, Lisa and Udell, David and Leech, Gavin and Mini, Ulisse and MacDiarmid, Monte},
  journal = {arXiv preprint arXiv:2308.10248},
  year    = {2023}
}

@article{zou2023representation,
  title   = {Representation Engineering: A Top-Down Approach to AI Transparency},
  author  = {Zou, Andy and Wang, Zifan and Carlini, Nicholas and Nasr, Milad and Kolter, J. Zico and Fredrikson, Matt},
  journal = {arXiv preprint arXiv:2310.01405},
  year    = {2023}
}

@misc{venhoff2025base,
  title  = {Base Models Know How to Reason, Thinking Models Learn When},
  author = {Venhoff, Constantin and Arcuschin, Iv{\'a}n and Torr, Philip and Conmy, Arthur and Nanda, Neel},
  note   = {Under review as a conference paper at ICLR 2026; arXiv:2510.07364},
  year   = {2025}
}

@article{kojima2022large,
  title   = {Large Language Models are Zero-Shot Reasoners},
  author  = {Kojima, Takeshi and Gu, Shixiang Shane and Reid, Machel and Matsuo, Yutaka and Iwasawa, Yutaka},
  journal = {arXiv preprint arXiv:2205.11916},
  year    = {2022}
}

@inproceedings{venhoff2025steering,
  title     = {Understanding Reasoning in Thinking Language Models via Steering Vectors},
  author    = {Venhoff, Constantin and Arcuschin, Iv{\'a}n and Torr, Philip and Conmy, Arthur and Nanda, Neel},
  booktitle = {Workshop on Reasoning and Planning for Large Language Models},
  year      = {2025}
}

@article{cunningham2023sparse,
  title   = {Sparse Autoencoders Find Highly Interpretable Features in Language Models},
  author  = {Cunningham, Hoagy and Ewart, Aidan and Riggs, Logan and Huben, Robert and Sharkey, Lee},
  journal = {arXiv preprint arXiv:2309.08600},
  year    = {2023}
}

@misc{templeton2024scaling,
  title  = {Scaling Monosemanticity: Extracting Interpretable Features from Claude 3 Sonnet},
  author = {Templeton, Adly and Conerly, Tom and Marcus, Jonathan and Lindsey, Jack and Bricken, Trenton and Chen, Brian and Pearce, Adam and Citro, Craig and Ameisen, Emmanuel and Jones, Andy and Cunningham, Hoagy and Turner, Nicholas L. and McDougall, Callum and MacDiarmid, Monte and Freeman, C. Daniel and Sumers, Theodore R. and Rees, Edward and Batson, Joshua and Jermyn, Adam and Carter, Shan and Olah, Chris and Henighan, Tom},
  howpublished = {Transformer Circuits Thread},
  year   = {2024},
  note   = {Online}
}

@article{zhang2025understanding,
  title={Understanding and Steering the Cognitive Behaviors of Reasoning Models at Test-Time},
  author={Zhang, Zhenyu and Wu, Xiaoxia and Zhou, Zhongzhu and Wu, Qingyang and Zhang, Yineng and Ponnusamy, Pragaash and Subbaraj, Harikaran and Wang, Jue and Song, Shuaiwen Leon and Athiwaratkun, Ben},
  journal={arXiv preprint arXiv:2512.24574},
  year={2025}
}

@article{ward2025reasoning,
  title={Reasoning-finetuning repurposes latent representations in base models},
  author={Ward, Jake and Lin, Chuqiao and Venhoff, Constantin and Nanda, Neel},
  journal={arXiv preprint arXiv:2507.12638},
  year={2025}
}

@article{huang2023large,
  title={Large language models cannot self-correct reasoning yet},
  author={Huang, Jie and Chen, Xinyun and Mishra, Swaroop and Zheng, Huaixiu Steven and Yu, Adams Wei and Song, Xinying and Zhou, Denny},
  journal={arXiv preprint arXiv:2310.01798},
  year={2023}
}

@article{gan2025rethinking,
  title={Rethinking external slow-thinking: From snowball errors to probability of correct reasoning},
  author={Gan, Zeyu and Liao, Yun and Liu, Yong},
  journal={arXiv preprint arXiv:2501.15602},
  year={2025}
}

@inproceedings{tyen2024llms,
  title={LLMs cannot find reasoning errors, but can correct them given the error location},
  author={Tyen, Gladys and Mansoor, Hassan and C{\u{a}}rbune, Victor and Chen, Yuanzhu Peter and Mak, Tony},
  booktitle={Findings of the Association for Computational Linguistics: ACL 2024},
  pages={13894--13908},
  year={2024}
}

@article{hendrycks2021measuring,
  title={Measuring mathematical problem solving with the math dataset},
  author={Hendrycks, Dan and Burns, Collin and Kadavath, Saurav and Arora, Akul and Basart, Steven and Tang, Eric and Song, Dawn and Steinhardt, Jacob},
  journal={arXiv preprint arXiv:2103.03874},
  year={2021}
}

@article{sheng2026solving,
  title={Solving inequality proofs with large language models},
  author={Sheng, Jiayi and Lyu, Luna and Jin, Jikai and Xia, Tanglin and Gu, Alex and Zou, James and Lu, Pan},
  journal={Advances in Neural Information Processing Systems},
  volume={38},
  year={2026}
}

@article{rein2023gpqa,
  title={Gpqa: A graduate-level google-proof q\&a benchmark},
  author={Rein, David and Hou, Betty Li and Stickland, Asa Cooper and Petty, Jackson and Pang, Richard Yuanzhe and Dirani, Julien and Michael, Julian and Bowman, Samuel R},
  journal={arXiv preprint arXiv:2311.12022},
  year={2023}
}

@article{zhou2022least,
  title={Least-to-most prompting enables complex reasoning in large language models},
  author={Zhou, Denny and Sch{\"a}rli, Nathanael and Hou, Le and Wei, Jason and Scales, Nathan and Wang, Xuezhi and Schuurmans, Dale and Cui, Claire and Bousquet, Olivier and Le, Quoc and others},
  journal={arXiv preprint arXiv:2205.10625},
  year={2022}
}

@inproceedings{zheng2024take,
  title={Take a step back: Evoking reasoning via abstraction in large language models},
  author={Zheng, Huaixiu Steven and Mishra, Swaroop and Chen, Xinyun and Cheng, Heng-Tze and Chi, Ed H and Le, Quoc V and Zhou, Denny},
  booktitle={International Conference on Learning Representations},
  volume={2024},
  pages={20279--20316},
  year={2024}
}

@inproceedings{weng2023large,
  title={Large language models are better reasoners with self-verification},
  author={Weng, Yixuan and Zhu, Minjun and Xia, Fei and Li, Bin and He, Shizhu and Liu, Shengping and Sun, Bin and Liu, Kang and Zhao, Jun},
  booktitle={Findings of the Association for Computational Linguistics: EMNLP 2023},
  pages={2550--2575},
  year={2023}
}

@inproceedings{wang2023plan,
  title={Plan-and-solve prompting: Improving zero-shot chain-of-thought reasoning by large language models},
  author={Wang, Lei and Xu, Wanyu and Lan, Yihuai and Hu, Zhiqiang and Lan, Yunshi and Lee, Roy Ka-Wei and Lim, Ee-Peng},
  booktitle={Proceedings of the 61st annual meeting of the association for computational linguistics (volume 1: long papers)},
  pages={2609--2634},
  year={2023}
}

@article{miao2023selfcheck,
  title={Selfcheck: Using llms to zero-shot check their own step-by-step reasoning},
  author={Miao, Ning and Teh, Yee Whye and Rainforth, Tom},
  journal={arXiv preprint arXiv:2308.00436},
  year={2023}
}

@inproceedings{dhuliawala2024chain,
  title={Chain-of-verification reduces hallucination in large language models},
  author={Dhuliawala, Shehzaad and Komeili, Mojtaba and Xu, Jing and Raileanu, Roberta and Li, Xian and Celikyilmaz, Asli and Weston, Jason},
  booktitle={Findings of the association for computational linguistics: ACL 2024},
  pages={3563--3578},
  year={2024}
}

@article{du2025latent,
  title={Latent thinking optimization: Your latent reasoning language model secretly encodes reward signals in its latent thoughts},
  author={Du, Hanwen and Dong, Yuxin and Ning, Xia},
  journal={arXiv preprint arXiv:2509.26314},
  year={2025}
}

@inproceedings{brown2020,
 author = {Brown, Tom and Mann, Benjamin and Ryder, Nick and Subbiah, Melanie and Kaplan, Jared D and Dhariwal, Prafulla and Neelakantan, Arvind and Shyam, Pranav and Sastry, Girish and Askell, Amanda and Agarwal, Sandhini and Herbert-Voss, Ariel and Krueger, Gretchen and Henighan, Tom and Child, Rewon and Ramesh, Aditya and Ziegler, Daniel and Wu, Jeffrey and Winter, Clemens and Hesse, Chris and Chen, Mark and Sigler, Eric and Litwin, Mateusz and Gray, Scott and Chess, Benjamin and Clark, Jack and Berner, Christopher and McCandlish, Sam and Radford, Alec and Sutskever, Ilya and Amodei, Dario},
 booktitle = {Advances in Neural Information Processing Systems},
 editor = {H. Larochelle and M. Ranzato and R. Hadsell and M.F. Balcan and H. Lin},
 pages = {1877--1901},
 publisher = {Curran Associates, Inc.},
 title = {Language Models are Few-Shot Learners},
 url = {https://proceedings.neurips.cc/paper_files/paper/2020/file/1457c0d6bfcb4967418bfb8ac142f64a-Paper.pdf},
 volume = {33},
 year = {2020}
}

@article{wang2026beyond,
  title={Beyond Dense States: Elevating Sparse Transcoders to Active Operators for Latent Reasoning},
  author={Wang, Yadong and Chen, Haodong and Tian, Yu and Geng, Chuanxing and Liang, Dong and Chen, Xiang},
  journal={arXiv preprint arXiv:2602.01695},
  year={2026}
}

@article{hu2026open,
  title={Open-reasoner-zero: An open source approach to scaling up reinforcement learning on the base model},
  author={Hu, Jingcheng and Zhang, Yinmin and Han, Qi and Jiang, Daxin and Zhang, Xiangyu and Shum, Heung-Yeung},
  journal={Advances in Neural Information Processing Systems},
  volume={38},
  pages={162239--162262},
  year={2026}
}

@misc{aime2024_i,
  title        = {2024 AIME I Problems and Solutions},
  author       = {{Art of Problem Solving}},
  year         = {2024},
  howpublished = {\url{https://artofproblemsolving.com/wiki/index.php/2024_AIME_I}},
  note         = {Accessed: 2026-05-24}
}

@misc{aime2024_ii,
  title        = {2024 AIME II Problems and Solutions},
  author       = {{Art of Problem Solving}},
  year         = {2024},
  howpublished = {\url{https://artofproblemsolving.com/wiki/index.php/2024_AIME_II}},
  note         = {Accessed: 2026-05-24}
}

@misc{aime2025_i,
  title        = {2025 AIME I Problems and Solutions},
  author       = {{Art of Problem Solving}},
  year         = {2025},
  howpublished = {\url{https://artofproblemsolving.com/wiki/index.php/2025_AIME_I}},
  note         = {Accessed: 2026-05-24}
}

@misc{aime2025_ii,
  title        = {2025 AIME II Problems and Solutions},
  author       = {{Art of Problem Solving}},
  year         = {2025},
  howpublished = {\url{https://artofproblemsolving.com/wiki/index.php/2025_AIME_II}},
  note         = {Accessed: 2026-05-24}
}

@misc{amc23_hf,
  title        = {AMC 2023 Dataset},
  author       = {{math-ai}},
  year         = {2025},
  publisher    = {Hugging Face},
  howpublished = {\url{https://huggingface.co/datasets/math-ai/amc23}},
  note         = {Accessed: 202-05-24}
}

@inproceedings{jin2025apeer,
  title={Apeer: Automatic prompt engineering enhances large language model reranking},
  author={Jin, Can and Peng, Hongwu and Zhao, Shiyu and Wang, Zhenting and Xu, Wujiang and Han, Ligong and Zhao, Jiahui and Zhong, Kai and Rajasekaran, Sanguthevar and Metaxas, Dimitris N},
  booktitle={Companion Proceedings of the ACM on Web Conference 2025},
  pages={2494--2502},
  year={2025}
}

@inproceedings{
jin2025two,
title={Two Heads are Better Than One: Test-time Scaling of Multi-agent Collaborative Reasoning},
author={Can Jin and Hongwu Peng and Qixin Zhang and Yujin Tang and Tong Che and Dimitris N. Metaxas},
booktitle={Workshop on Scaling Environments for Agents},
year={2025},
url={https://openreview.net/forum?id=aLGgp4FK0A}
}

@article{jin2026reasoning,
  title={Reasoning over Precedents Alongside Statutes: Case-Augmented Deliberative Alignment for LLM Safety},
  author={Jin, Can and Wu, Rui and Che, Tong and Zhang, Qixin and Peng, Hongwu and Zhao, Jiahui and Wang, Zhenting and Wei, Wenqi and Han, Ligong and Zhang, Zhao and others},
  journal={arXiv preprint arXiv:2601.08000},
  year={2026}
}

@article{jin2025your,
  title={Your reward function for rl is your best prm for search: Unifying rl and search-based tts},
  author={Jin, Can and Zhou, Yang and Zhang, Qixin and Peng, Hongwu and Zhang, Di and Dong, Zihan and Pavone, Marco and Han, Ligong and Hong, Zhang-Wei and Che, Tong and others},
  journal={arXiv preprint arXiv:2508.14313},
  year={2025}
}

@article{zhang2026soft,
  title={Soft thinking: Unlocking the reasoning potential of llms in continuous concept space},
  author={Zhang, Zhen and He, Xuehai and Yan, Weixiang and Shen, Ao and Zhao, Chenyang and Wang, Xin},
  journal={Advances in Neural Information Processing Systems},
  volume={38},
  pages={168990--169012},
  year={2026}
}

@article{zhang2026cm2,
  title={CM2: Reinforcement Learning with Checklist Rewards for Multi-Turn and Multi-Step Agentic Tool Use},
  author={Zhang, Zhen and Song, Kaiqiang and Wang, Xun and Hu, Yebowen and Yan, Weixiang and Zhao, Chenyang and Zou, Henry Peng and Deng, Haoyun and Indurthi, Sathish Reddy and Liu, Shujian and others},
  journal={arXiv preprint arXiv:2602.12268},
  year={2026}
}

@article{zhang2026locate,
  title={Locate, steer, and improve: A practical survey of actionable mechanistic interpretability in large language models},
  author={Zhang, Hengyuan and Zhang, Zhihao and Wang, Mingyang and Su, Zunhai and Wang, Yiwei and Wang, Qianli and Yuan, Shuzhou and Nie, Ercong and Duan, Xufeng and Han, Feijiang and others},
  journal={arXiv preprint arXiv:2601.14004},
  year={2026}
}
\newpage
\appendix

\section{Training and Experimental Details}
\label{app:training_exp_details}

\paragraph{Models and intervention layer.}
We use \textsc{Open-Reasoner-7B} as the primary model and \textsc{Open-Reasoner-1.5B} as a smaller-model comparison within the same family. The base model parameters are never updated.

\paragraph{Zero-shot decoding and scoring.}
Base, \ours Basic, and full \ours are evaluated in a zero-shot setting with greedy decoding and batch size $1$. In these configurations, the model receives only the original problem statement, without few-shot demonstrations or task-specific initial reasoning prompts. The CoT and five-shot baselines use the same decoding settings with the prompts described below. Final answers are parsed from the full output and scored with the benchmark-specific evaluator: math-style benchmarks use rule-based or math-verification scoring where applicable, while multiple-choice tasks use extracted option letters against the gold answer.

\paragraph{Prompt baselines.}
The CoT baseline uses the same problem statement with a fixed instruction prefix:
\begin{figure}[t]
\begin{AIBox}[CoT Prompt]
\small
Let's solve this problem step by step, problem: \texttt{<problem>}
\end{AIBox}
\caption{Zero-shot chain-of-thought prompt used as a baseline.}
\label{fig:cot_prompt}
\end{figure}
For few-shot prompting, we use fixed five-example prompts before the test problem. Math-style datasets use elementary algebra, LCM, geometry, counting, and summation examples. \textsc{GPQA-Diamond} uses five multiple-choice science examples and \textsc{IneqMath} uses five inequality examples covering AM-GM, triangle inequality, and algebraic nonnegativity. The full prompt templates are shown below.

\begin{figure*}[p]
\begin{AIBox}[Five-Shot Math Prompt]
\small
Here are 5 example problems and their solutions.

\textbf{Example 1.} Problem: If $3x + 2 = 11$, what is the value of $x^2$? Solution: From $3x + 2 = 11$ we get $3x = 9$, so $x = 3$. Therefore $x^2 = 9$. Final answer: $\boxed{9}$

\textbf{Example 2.} Problem: What is the smallest positive integer divisible by both 6 and 8? Solution: Prime factorize: $6 = 2 \cdot 3$ and $8 = 2^3$. So $\mathrm{lcm}(6,8) = 2^3 \cdot 3 = 24$. Final answer: $\boxed{24}$

\textbf{Example 3.} Problem: A right triangle has legs of length 5 and 12. What is the length of the hypotenuse? Solution: By the Pythagorean theorem, $c = \sqrt{5^2 + 12^2} = \sqrt{169} = 13$. Final answer: $\boxed{13}$

\textbf{Example 4.} Problem: How many ways are there to choose 3 books from a shelf of 7 distinct books? Solution: This is $\binom{7}{3} = \frac{7 \cdot 6 \cdot 5}{3 \cdot 2 \cdot 1} = 35$. Final answer: $\boxed{35}$

\textbf{Example 5.} Problem: What is the sum of the first 100 positive integers? Solution: Using $\sum_{k=1}^{n} k = \frac{n(n+1)}{2}$ with $n = 100$, the sum is $\frac{100 \cdot 101}{2} = 5050$. Final answer: $\boxed{5050}$

Now solve this problem.

Problem: \texttt{<problem>}
\end{AIBox}
\caption{Five-shot prompt template for math-style benchmarks.}
\label{fig:five_shot_math_prompt}
\end{figure*}

\begin{figure*}[p]
\begin{AIBox}[Five-Shot GPQA Prompt]
\small
Here are 5 example multiple-choice problems and their solutions.

\textbf{Example 1.} Problem: A ball is dropped from a height of 20 m. How long does it take to hit the ground? Ignore air resistance, $g = 10\,\mathrm{m/s^2}$. Choices: (A) 1 s, (B) 2 s, (C) 4 s, (D) 5 s. Solution: Using $h = \frac{1}{2}gt^2$, solve $20 = 5t^2$, so $t=2$. Final answer: $\boxed{B}$

\textbf{Example 2.} Problem: Which of the following has the highest electronegativity? Choices: (A) Carbon, (B) Nitrogen, (C) Oxygen, (D) Fluorine. Solution: Electronegativity increases across a period and decreases down a group. Fluorine has the highest value. Final answer: $\boxed{D}$

\textbf{Example 3.} Problem: Which organelle is responsible for ATP production via oxidative phosphorylation in eukaryotic cells? Choices: (A) Nucleus, (B) Ribosome, (C) Mitochondrion, (D) Endoplasmic reticulum. Solution: The mitochondrion houses the electron transport chain and ATP synthase. Final answer: $\boxed{C}$

\textbf{Example 4.} Problem: A 2 kg object moves at 3 m/s. What is its kinetic energy? Choices: (A) 3 J, (B) 6 J, (C) 9 J, (D) 12 J. Solution: $KE=\frac{1}{2}mv^2=\frac{1}{2}(2)(3)^2=9\,\mathrm{J}$. Final answer: $\boxed{C}$

\textbf{Example 5.} Problem: What is the pH of a $10^{-3}$ M HCl solution? Choices: (A) 1, (B) 3, (C) 7, (D) 11. Solution: HCl fully dissociates, so $[\mathrm{H}^+]=10^{-3}$ M and $\mathrm{pH}=3$. Final answer: $\boxed{B}$

Now solve this problem.

Problem: \texttt{<problem>}
\end{AIBox}
\caption{Five-shot prompt template for GPQA-Diamond.}
\label{fig:five_shot_gpqa_prompt}
\end{figure*}

\begin{figure*}[p]
\begin{AIBox}[Five-Shot IneqMath Prompt]
\small
Here are 5 example inequality problems and their solutions.

\textbf{Example 1.} Problem: Find the largest constant $C$ such that $x^2+y^2 \geq Cxy$ for all real $x,y$. Solution: $x^2+y^2-2xy=(x-y)^2\geq0$, so $C=2$ is tight. Final answer: $\boxed{2}$

\textbf{Example 2.} Problem: For all positive reals $a,b$, determine the relation between $a+b$ and $2\sqrt{ab}$. Solution: By AM-GM, $\frac{a+b}{2}\geq\sqrt{ab}$, so $a+b\geq2\sqrt{ab}$. Final answer: $\boxed{\geq}$

\textbf{Example 3.} Problem: Find the smallest constant $C$ such that $|x+y|\leq C(|x|+|y|)$ for all real $x,y$. Solution: By the triangle inequality, $C=1$ works and is tight. Final answer: $\boxed{1}$

\textbf{Example 4.} Problem: For positive reals $a,b,c$, determine the relation between $a^2+b^2+c^2$ and $ab+bc+ca$. Solution: $a^2+b^2+c^2-ab-bc-ca=\frac{1}{2}((a-b)^2+(b-c)^2+(c-a)^2)\geq0$. Final answer: $\boxed{\geq}$

\textbf{Example 5.} Problem: Find the largest constant $C$ such that $(a+b)^2\geq Cab$ for all positive reals $a,b$. Solution: By AM-GM, $a+b\geq2\sqrt{ab}$, so $(a+b)^2\geq4ab$, with equality at $a=b$. Final answer: $\boxed{4}$

Now solve this problem.

Problem: \texttt{<problem>}
\end{AIBox}
\caption{Five-shot prompt template for IneqMath.}
\label{fig:five_shot_ineqmath_prompt}
\end{figure*}

\paragraph{Reward model training.}
The reward model is trained on sparse latent reasoning sequences obtained by encoding the model's hidden activations through the pretrained SAE (Section~\ref{sec:problem-setup}). For each generation, we keep only the reasoning segment starting from the model's ``think'' marker, discarding the prompt prefix, and pair the sequence with a binary final-answer correctness label that is repeated over all positions for token-level supervision. Architecturally, the reward model is a lightweight Transformer encoder operating directly on $d_z$-dimensional SAE latents: it applies a LayerNorm on the raw SAE input (essential, since unnormalized SAE activations are often near-zero and easily dominated by positional encoding), a linear projection to hidden size $d=128$, sinusoidal positional encoding, two Transformer encoder blocks (4 attention heads, feedforward width $4d$, dropout $0.1$), and an MLP head (Linear--ReLU--Dropout--Linear--Sigmoid) producing a per-position probability $p_{i,t} \in (0,1)$. We optimize with AdamW (learning rate $5\times 10^{-4}$, weight decay $10^{-4}$), binary cross-entropy loss, gradient clipping at max-norm $1.0$, and batch size $1$ (one variable-length trace per step). To counter class imbalance between correct and incorrect trajectories, we apply weighted random sampling, with each trace sampled with probability inversely proportional to its class frequency. We train for $30$ epochs and select the checkpoint with the lowest training loss for inference-time steering. No behavior labels, GPT-4o annotations, or predefined cognitive categories are used at any stage and the only supervision is final-answer correctness.

\paragraph{Steering variants and gate.}
\textsc{\ours(basic)} applies reward-guided latent steering without the reward and confidence gate. Full \textsc{\ours} uses the gate to selectively intervene: steering is triggered when the reward score is below the reward threshold, or when the reward score is above threshold but the previous-token decoding confidence, measured as the last-token maximum probability, is below the confidence threshold. In the experiments below, most conditions are instantiated as reward $<0.9$ or reward $\ge0.9$ with last-token maximum probability $<0.72$, except for configurations whose thresholds are listed separately in Table~\ref{tab:appendix_exp_config_model}.

\begin{table*}[t]
    \centering
    \renewcommand{\arraystretch}{1.1}
    \small
    \begin{tabular}{llccccc}
        \toprule
        Dataset & Model & $K$ & $\alpha$ & Reward & Confidence & Device \\
        \midrule
        AMC23        & Open-Reasoner-7B   & 1 & 1.400 & 0.9 & 0.72 & RTX A4500 \\
        AMC23        & Open-Reasoner-1.5B & 4 & 0.300 & 0.9 & 0.72 & RTX A6000 \\
        \midrule
        AIME24       & Open-Reasoner-7B   & 2 & 0.295 & 0.9 & 0.72 & RTX A4500 \\
        AIME24       & Open-Reasoner-1.5B & 4 & 0.900 & 0.9 & 0.72 & RTX A4500 \\
        \midrule
        AIME25       & Open-Reasoner-7B   & 3 & 1.320 & 0.9 & 0.72 & RTX A4500 \\
        AIME25       & Open-Reasoner-1.5B & 2 & 0.400 & 0.9 & 0.72 & RTX A6000 \\
        \midrule
        GPQA Diamond & Open-Reasoner-7B   & 4 & 1.150 & 0.9 & 0.72 & RTX A5000 \\
        GPQA Diamond & Open-Reasoner-1.5B & 3 & 1.000 & 0.9 & 0.72 & RTX A5000 \\
        \midrule
        IneqMath     & Open-Reasoner-7B   & 2 & 0.700 & 0.9 & 0.72 & RTX A5000 \\
        IneqMath     & Open-Reasoner-1.5B & 2 & 1.100 & 0.9 & 0.72 & RTX A6000 \\
        \midrule
        MATH-500     & Open-Reasoner-7B   & 1 & 1.400 & 0.8 & 0.69 & RTX A4500 \\
        MATH-500     & Open-Reasoner-1.5B & 1 & 0.100 & 0.9 & 0.72 & RTX A6000 \\
        \bottomrule
    \end{tabular}
    \caption{Experimental configurations for different datasets and models. $K$ denotes the number of latent optimization steps, and $\alpha$ denotes the step size. The reward and confidence columns report the corresponding gate thresholds. The device column reports the GPU used for each dataset--model configuration.}
    \label{tab:appendix_exp_config_model}
    \vspace{0.3em}
    \begin{minipage}{0.92\textwidth}
        \small
    \end{minipage}
\end{table*}

\section{Behavior Annotation Protocol}
\label{app:behavior_annotation}

We use GPT-4o only for post-hoc diagnostic annotation of generated reasoning traces. The same fixed prompt is applied to base and \ours outputs. These annotations are never used for reward-model training, steering-vector construction, or gate triggering.

\paragraph{Annotation prompt.}
We apply the same fixed GPT-4o prompt to both base and \ours traces, using the behavior criteria below.

\begin{figure*}[p]
\begin{AIBox}[GPT-4o Behavior Annotation Prompt]
\small
\textbf{Input.} You will receive two fields: \texttt{problem}, the original problem statement, and \texttt{reasoning\_trace}, the generated solution.\\[0.25em]
\textbf{Task.} For each behavior below, decide whether it appears at least once anywhere in the reasoning trace. Do not judge final-answer correctness except when the trace explicitly verifies a result against the original problem.\\[0.25em]
\textbf{Output.} Return only a JSON object with one key per behavior. Each key must contain a Boolean field \texttt{present} and a short \texttt{evidence} string. If a behavior is absent, set \texttt{present} to \texttt{false} and use an empty evidence string.\\[0.25em]
\textbf{Schema.} \texttt{\{StrategicPlanning: \{present: bool, evidence: str\}, StructuredDecomposition: \{present: bool, evidence: str\}, ConstraintGrounding: \{present: bool, evidence: str\}, CourseCorrection: \{present: bool, evidence: str\}, SolutionVerification: \{present: bool, evidence: str\}\}}
\end{AIBox}
\caption{Prompt used for post-hoc GPT-4o behavior annotation.}
\label{fig:gpt4o_annotation_prompt}
\end{figure*}

\begin{itemize}[leftmargin=1.3em,itemsep=2pt,topsep=2pt]
    \item \textbf{Strategic Planning:} mark true if the trace selects a method, theorem, or overall strategy before detailed computation and explains why it is appropriate.
    \item \textbf{Structured Decomposition:} mark true if the trace breaks the problem into cases, subproblems, lemmas, branches, or explicitly named intermediate goals.
    \item \textbf{Constraint Grounding:} mark true if the trace actively uses problem constraints to check domains, exclude invalid solutions, validate boundary conditions, or restrict the search space.
    \item \textbf{Course Correction:} mark true if the trace detects an error, contradiction, missing case, or uncertainty and then revises the computation, switches methods, or redirects the solution path.
    \item \textbf{Solution Verification:} mark true if the trace substitutes an intermediate result or final answer back into the original problem, or otherwise checks that the derived answer satisfies the required conditions.
\end{itemize}

\paragraph{Occurrence and improvement.}
For each trace, a behavior has occurrence value $1$ if GPT-4o marks it as present at least once and $0$ otherwise. The occurrence rate is the average of this indicator over the matched trace set. Behavior improvement is computed as the \ours occurrence rate minus the Base occurrence rate. Because the labels are post-hoc diagnostics, they support process-level interpretation rather than direct claims that \ours explicitly controls predefined behaviors.

\begin{table*}[t]
\centering
\footnotesize
\setlength{\tabcolsep}{4pt}
\renewcommand{\arraystretch}{1.15}
\begin{tabularx}{\textwidth}{@{}>{\raggedright\arraybackslash}p{0.16\textwidth}>{\raggedright\arraybackslash}p{0.46\textwidth}>{\raggedright\arraybackslash}X@{}}
\toprule
\textbf{Behavior} & \textbf{Trace-level criterion} & \textbf{Associated SAE dimensions} \\
\midrule
Strategic Planning & Selects a method or theorem and explains the rationale before detailed computation, e.g., choosing a Diophantine strategy before algebraic execution. & $z_8$ meta-cognitive strategy and $z_1$ logical branching. \\
Structured Decomposition & Breaks the problem into independent branches, such as case splits, named lemmas, or formally defined intermediate sub-problems. & $z_1$ theorem invocation and $z_0$ structural decomposition. \\
Constraint Grounding & Uses problem constraints to guide or limit the solution process, such as checking domains, excluding invalid solutions, or verifying boundary conditions. & $z_7$ constraint processing and $z_6$ deductive conclusion. \\
Course Correction & Detects a problem during reasoning and adjusts direction, such as switching methods after a contradiction or revising a flawed assumption. & $z_8$ meta-cognitive re-planning and $z_7$ violation detection and $z_1$ logical branching. \\
Solution Verification & Substitutes the final answer back into the original problem or checks whether the derived result satisfies the required conditions. & $z_7$ constraint processing and $z_6$ answer consolidation. \\
\bottomrule
\end{tabularx}
\caption{Cognitive behavior categories used for matched base-vs.-\ours trace analysis. Associated SAE dimensions are used only as interpretability priors for analyzing $\Delta z$, not as deterministic behavior mappings.}
\label{tab:case_study_behavior}
\end{table*}

\section{Case Diagnostic Notes}
\label{app:additional_analyses}

This appendix clarifies how to read the diagnostics used in the qualitative cases. The case-study boxes report steering-event counts, early trigger positions, and the largest corrected SAE dimensions. These diagnostics are derived from per-question SAE steering traces and support a cautious process-level reading: they show where latent updates were applied and which sparse dimensions changed most, but they do not provide token-level reward values or prove a causal mechanism. We therefore avoid numeric reward claims and interpret dimension names only as post-hoc priors from Table~\ref{tab:sae_interpretability}.

\section{Qualitative Case Studies with Reward-Signal Diagnostics}
\label{app:case_studies}

We provide representative examples where \ours changes an initially incorrect single generation into a correct answer.
Each case highlights the critical divergence between the base and \ours-steered trajectories.
These examples cover different failure modes, including combinatorial overcounting, missing cases, incorrect constraint handling, scientific concept confusion, and an incorrect equality conclusion. The Q84 case also illustrates a correct final answer with a flawed proof.
The highlighted failure modes are used only for post-hoc analysis. \ours itself does not rely on behavior labels or predefined steering directions.
Excerpts are shortened for readability while preserving the displayed reasoning steps and final answers. Author analyses identify both corrected errors and proof errors that remain in the steered outputs.
SAE dimension interpretations are post-hoc priors and should not be read as deterministic labels, especially for non-mathematical tasks such as GPQA.

\begin{figure*}[t]
    \centering
\includegraphics[width=0.95\textwidth]{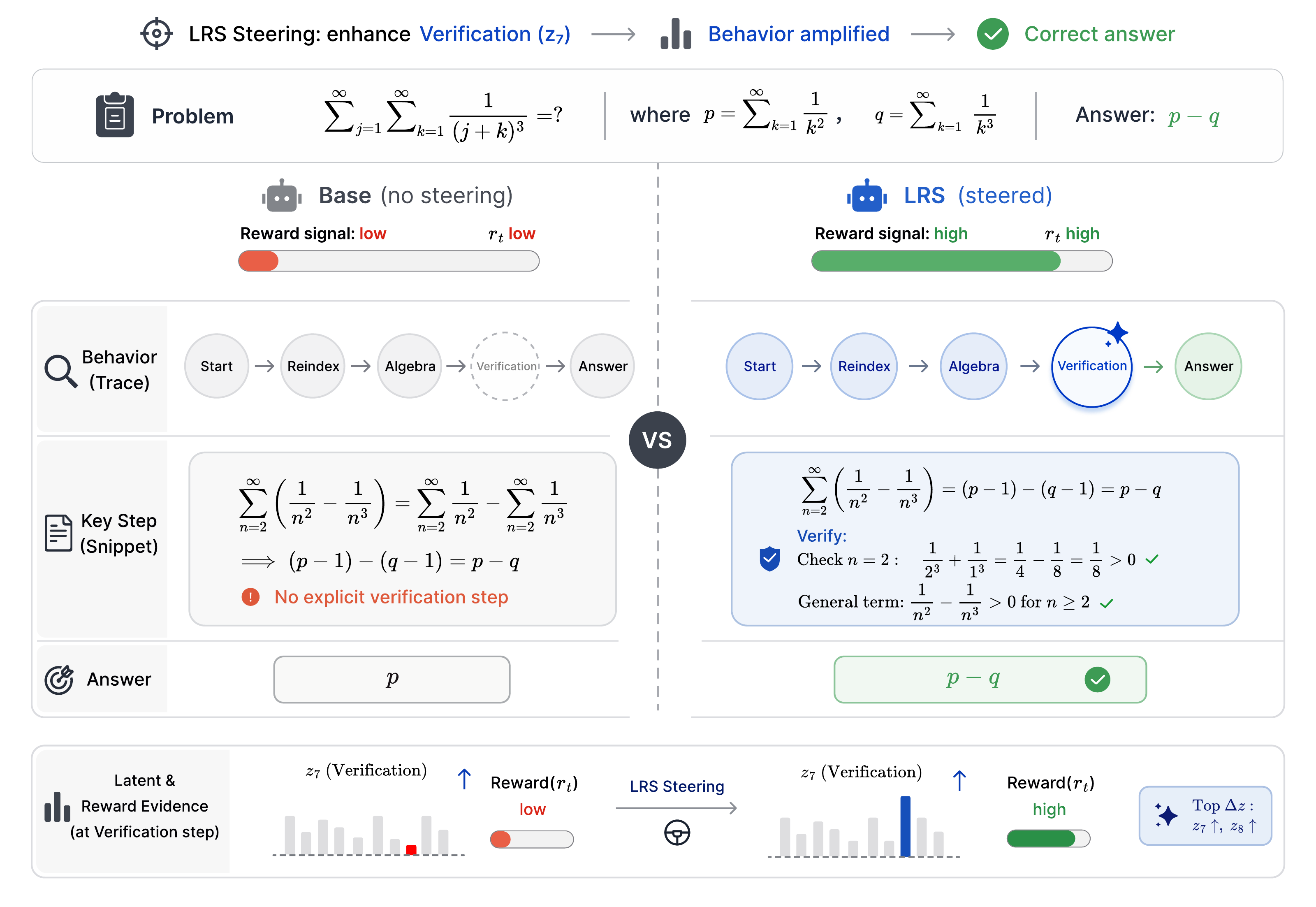}
    \caption{Qualitative example where \ours steers an incorrect reasoning trace toward verification and a correct answer.}
    \label{fig:lrs_verification_case}
\end{figure*}

\begin{figure*}[!t]
\begin{AIBox}[Case Study 1: AMC23 Q38 --- Combinatorics (Subset Counting)]
\parbox[t]{\textwidth}{
\small

\PromptSection{Question}
Determine the number of nonempty subsets $B$ of $\{0,1,2,\ldots,12\}$ such that the number of elements in $B$ is equal to the least element of $B$.

\noindent\textcolor{AIgrayline}{\rule{\linewidth}{0.35pt}}

\PromptSection{Base Output}
The base model lets $k$ denote the least element of $B$ and correctly requires $|B|=k$ with $k\ge1$.
It then chooses $k-1$ additional elements from $\{k, k+1, \ldots, 12\}$, which contains $13-k$ elements, yielding $\textcolor{BaseErrRed}{\binom{13-k}{k-1}}$.
Summing for $k=1$ to $7$: $1+11+45+84+70+21+1 = \textcolor{BaseErrRed}{233}$.
The error is that $k$ itself is already required to be in $B$, so the remaining $k-1$ elements must be chosen from $\{k{+}1,\ldots,12\}$, which has $12-k$ elements, not from $\{k,\ldots,12\}$.

\noindent\textcolor{AIgrayline}{\rule{\linewidth}{0.35pt}}

\PromptSection{\ours Steered Output}
The \ours-steered trace uses the correct selection pool: $\textcolor{SteerFixGreen}{\binom{12-k}{k-1}}$.
``The subset $B$ must include $k$ and $k{-}1$ other elements, all of which must be greater than $k$.
The elements of $B$ can be chosen from the set $\{k{+}1, k{+}2, \ldots, 12\}$.
The number of elements in this set is $12-k$.''
Summing for $k=1$ to $6$: $1+10+36+56+35+6 = \textcolor{SteerFixGreen}{144}$.

\noindent\textcolor{AIgrayline}{\rule{\linewidth}{0.35pt}}

\PromptSection{Key Difference}
\textcolor{BaseErrRed}{Base} uses $\binom{13-k}{k-1}$, which includes $k$ in the selection pool and causes overcounting.
\textcolor{SteerFixGreen}{\ours} uses $\binom{12-k}{k-1}$, correctly excluding $k$.

\PromptSection{Steering Info}
184 steering events and the first recorded steering events occur at generation steps 24, 59, and 69.
Top corrected dimensions: $z_2$ $\uparrow$ (algebraic execution), $z_4$ $\uparrow$ (variable extraction), $z_9$ $\downarrow$ (complexity evaluation).

\PromptSection{Reward-Signal Diagnostic}
The base reasoning trace begins to fail when it constructs the counting pool $\{k,\ldots,12\}$ and thereby counts the required minimum element $k$ twice.
The reward and confidence gate is associated with early interventions before the final summation, suggesting that the local latent state around the combinatorial setup was treated as fragile.
In the steered trace, the solution grounds the constraint that $k$ is already included in $B$ and changes the pool to $\{k+1,\ldots,12\}$.
This change is consistent with \emph{Constraint Grounding} and \emph{Course Correction}, rather than a behavior label explicitly supplied to \ours.

\PromptSection{Latent-Change Interpretation}
The largest average latent changes up-regulate $z_2$ and $z_4$, which Table~\ref{tab:sae_interpretability} associates with algebraic execution and variable initialization, while down-regulating $z_9$, associated with complexity evaluation.
This pattern is consistent with the steered trace correcting the counting setup and avoiding an overcomplicated selection pool.
These dimension names are post-hoc interpretability priors, not deterministic causal labels.
}
\end{AIBox}
\caption{AMC23 Q38: \ours corrects an overcount caused by including $k$ in the selection pool.}
\label{fig:case_amc23_q38}
\end{figure*}

\begin{figure*}[!t]
\begin{AIBox}[Case Study 2: AIME25 Q2 --- Counting (Multinomial Partitions)]
\parbox[t]{\textwidth}{
\small

\PromptSection{Question}
Nine players each choose one of three flavors: chocolate, vanilla, or strawberry.
Each flavor is chosen by at least one player, and the number choosing chocolate is strictly greater than the number choosing vanilla, which is strictly greater than the number choosing strawberry.
Find the number of valid assignments modulo $1000$.

\noindent\textcolor{AIgrayline}{\rule{\linewidth}{0.35pt}}

\PromptSection{Base Output}
Setting $c>v>s\ge1$ with $c+v+s=9$, the base model enumerates:
\begin{itemize}[leftmargin=1.2em,itemsep=1pt,topsep=2pt]
  \item $s=1$: finds $(5,3,1)$ as valid, but \textcolor{BaseErrRed}{rejects $(6,2,1)$ by incorrectly claiming $v>s$ fails}, although $2>1$ holds.
  \item $s=2$: tests $(4,3,2)$ but \textcolor{BaseErrRed}{rejects it, writing ``not valid since $c>v$''}, which misapplies the constraint.
\end{itemize}
Only the single partition $(5,3,1)$ is retained:
$\binom{9}{5,3,1}=504$, so the answer is $\textcolor{BaseErrRed}{504}$.

\noindent\textcolor{AIgrayline}{\rule{\linewidth}{0.35pt}}

\PromptSection{\ours Steered Output}
The \ours-steered reasoning trace correctly enumerates all three valid partitions:
\begin{itemize}[leftmargin=1.2em,itemsep=1pt,topsep=2pt]
  \item $(6,2,1)$: \textcolor{SteerFixGreen}{valid since $6>2>1$}, $\binom{9}{6,2,1}=252$.
  \item $(5,3,1)$: valid, $\binom{9}{5,3,1}=504$.
  \item $(4,3,2)$: \textcolor{SteerFixGreen}{valid since $4>3>2$}, $\binom{9}{4,3,2}=1260$.
\end{itemize}
Total: $252+504+1260=2016$.
Remainder modulo $1000$: $\textcolor{SteerFixGreen}{16}$.

\noindent\textcolor{AIgrayline}{\rule{\linewidth}{0.35pt}}

\PromptSection{Key Difference}
\textcolor{BaseErrRed}{Base} misses two of three valid partitions due to incorrect constraint checks.
\textcolor{SteerFixGreen}{\ours} correctly enumerates all three.

\PromptSection{Steering Info}
134 steering events and the first recorded steering events occur at generation steps 14, 15, and 18.
Top corrected dimensions: $z_2$ $\uparrow$ (algebraic execution), $z_3$ $\uparrow$ (symbolic reasoning), $z_0$ $\downarrow$ (geometric modeling).

\PromptSection{Reward-Signal Diagnostic}
The base reasoning trace starts to fail during case enumeration: it rejects valid partitions such as $(6,2,1)$ and $(4,3,2)$ despite the strict ordering constraints being satisfied.
The early reward and confidence gate activations suggest that the latent state during the enumeration phase was flagged as locally fragile.
The steered trace revisits the missing cases and evaluates all valid multinomial partitions before applying the modulo operation.
This pattern is associated with \emph{Structured Decomposition}, \emph{Constraint Grounding}, and \emph{Course Correction}.

\PromptSection{Latent-Change Interpretation}
The dominant changes up-regulate $z_2$ and $z_3$, associated with algebraic execution and symbolic formatting, while down-regulating $z_0$, associated with structural modeling.
This is consistent with the steered trace moving from an incomplete structural enumeration to explicit validation of each multinomial case.
These dimension names are post-hoc interpretability priors, not deterministic causal labels.
}
\end{AIBox}
\caption{AIME25 Q2: \ours recovers the missed valid partitions and the correct remainder.}
\label{fig:case_aime25_q2}
\end{figure*}

\begin{figure*}[!t]
\begin{AIBox}[Case Study 3: AMC23 Q25 --- Complex Number Algebra]
\parbox[t]{\textwidth}{
\small

\PromptSection{Question}
Define the operation $u \otimes v = ac + bdi$, where $u=a+bi$ and $v=c+di$.
Find $|z|^2$ given that $z \otimes z = z^2 + 40$.

\noindent\textcolor{AIgrayline}{\rule{\linewidth}{0.35pt}}

\PromptSection{Base Output}
Both models set $z=x+yi$ and correctly derive the real-part constraint $y^2=40$.
For the imaginary part, the base model incorrectly writes $\textcolor{BaseErrRed}{y^2 i = 2xy}$, omitting the factor $i$ on the right-hand side.
The base model argues:
``The left side $y^2 i$ is purely imaginary, and the right side $2xy$ is a real number.
\textcolor{BaseErrRed}{For these to be equal \ldots\ this implies $2xy=0$ and $y^2=0$.}
However, we already found that $y^2=40$, which means $y\neq0$.
Therefore, the only way \ldots\ is if $x=0$.''
With $x=0$ and $y^2=40$, it concludes $|z|^2 = 0+40 = \textcolor{BaseErrRed}{40}$.

\noindent\textcolor{AIgrayline}{\rule{\linewidth}{0.35pt}}

\PromptSection{\ours Steered Output}
The \ours-steered reasoning trace instead obtains the imaginary-coefficient relation:
``\textcolor{SteerFixGreen}{This implies $y^2 = 2xy$.}
If $y\neq0$, we can divide both sides by $y$:
\textcolor{SteerFixGreen}{$y = 2x$.}''
Combined with $y^2=40$, this gives $(2x)^2=40 \Rightarrow x^2=10$.
Therefore, $|z|^2 = x^2+y^2 = 10+40 = \textcolor{SteerFixGreen}{50}$.

\noindent\textcolor{AIgrayline}{\rule{\linewidth}{0.35pt}}

\PromptSection{Key Difference}
\textcolor{BaseErrRed}{Base} omits the factor $i$ from $2xyi$, then reasons from the incorrect equation $y^2 i = 2xy$ and concludes $x=0$ despite the contradiction with $y^2=40$.
\textcolor{SteerFixGreen}{\ours} correctly compares the imaginary coefficients to obtain $y=2x$.

\PromptSection{Mathematical Check (Author Analysis)}
The original equation expands to
\[
x^2+y^2i=(x^2-y^2)+2xyi+40.
\]
Comparing real and imaginary parts gives $y^2=40$ and $y^2i=2xyi$, respectively. Thus $y^2=2xy$. Since $y\neq0$, we obtain $y=2x$, $x^2=10$, and $|z|^2=50$.

\PromptSection{Steering Info}
247 steering events and the first recorded steering events occur at generation steps 9, 37, and 48.
Top corrected dimensions: $z_2$ $\uparrow$ (algebraic execution), $z_9$ $\downarrow$ (complexity evaluation), $z_4$ $\uparrow$ (variable extraction).

\PromptSection{Reward-Signal Diagnostic}
The base reasoning trace starts to fail when it separates the imaginary-part equation incorrectly and forces $x=0$ instead of comparing imaginary coefficients.
The reward and confidence gate is associated with interventions near the algebraic manipulation stage, suggesting low reliability for the local equation-handling state.
In the steered trace, the solution compares imaginary coefficients directly, derives $y=2x$, and then completes the norm computation.
This shift is consistent with \emph{Constraint Grounding} and \emph{Course Correction}.

\PromptSection{Latent-Change Interpretation}
The largest average changes up-regulate $z_2$ and $z_4$, associated with algebraic execution and variable extraction, while down-regulating $z_9$, associated with complexity evaluation.
This pattern is consistent with the steered trace replacing an invalid coefficient split with direct manipulation of the real and imaginary constraints.
These dimension names are post-hoc interpretability priors, not deterministic causal labels.
}
\end{AIBox}
\caption{AMC23 Q25: \ours fixes the imaginary-part constraint and obtains $|z|^2=50$.}
\label{fig:case_amc23_q25}
\end{figure*}

\begin{figure*}[!t]
\begin{AIBox}[Case Study 4: GPQA Diamond Q39 --- Chemistry (Separation Science)]
\parbox[t]{\textwidth}{
\small

\PromptSection{Question}
A synthetic organic chemist tells a colleague: ``My compounds are on top of each other.''
What is the second chemist most likely referring to?
\begin{itemize}[leftmargin=1.2em,itemsep=1pt,topsep=2pt]
  \item[(A)] The compounds have similar polarities.
  \item[(B)] The compounds are bonding through non-covalent interactions.
  \item[(C)] The compounds have similar optical rotations.
  \item[(D)] The compounds have similar boiling points.
\end{itemize}

\noindent\textcolor{AIgrayline}{\rule{\linewidth}{0.35pt}}

\PromptSection{Base Output}
The base model considers both chromatography and distillation but ultimately favors distillation:
``The most likely reason for the compounds being `on top of each other' \ldots\ is that they have \textcolor{BaseErrRed}{similar boiling points}.
This would make it difficult to separate them using distillation.''
Answer: \textcolor{BaseErrRed}{D}.

\noindent\textcolor{AIgrayline}{\rule{\linewidth}{0.35pt}}

\PromptSection{\ours Steered Output}
The \ours-steered trace identifies the chromatography context:
``\textcolor{SteerFixGreen}{Similar polarities (option A)} is a broad property that can affect the solubility and separation of compounds, making it a plausible explanation for the observed issue.
The phrase `on top of each other' suggests that the compounds are not well-separated or are difficult to distinguish from one another.''
Answer: \textcolor{SteerFixGreen}{A}.

\noindent\textcolor{AIgrayline}{\rule{\linewidth}{0.35pt}}

\PromptSection{Key Difference}
\textcolor{BaseErrRed}{Base} interprets ``on top of each other'' through the lens of distillation and boiling points.
\textcolor{SteerFixGreen}{\ours} connects the phrase to chromatographic separation and polarities.

\PromptSection{Steering Info}
653 steering events and the first recorded steering events occur at generation steps 2, 3, and 6.
Top corrected dimensions: $z_3$ $\uparrow$ (symbolic reasoning), $z_2$ $\uparrow$ (algebraic execution), $z_0$ $\downarrow$ (geometric modeling).

\PromptSection{Reward-Signal Diagnostic}
The base reasoning trace begins to drift when it interprets ``on top of each other'' through a distillation frame and selects boiling points, rather than grounding the phrase in chromatographic separation.
The very early gate activations suggest that the reward and confidence signal treated the initial conceptual framing as unstable.
The steered trace redirects the interpretation toward compounds co-eluting or poorly separating on a chromatographic medium, leading to similar polarities.
This change is associated with \emph{Strategic Planning} at the conceptual-framing level and \emph{Constraint Grounding} in the domain-specific clue.

\PromptSection{Latent-Change Interpretation}
The largest average changes up-regulate $z_3$ and $z_2$ and down-regulate $z_0$.
Although Table~\ref{tab:sae_interpretability} names these dimensions using math-heavy max-activating contexts, for this non-mathematical GPQA case they should be read only as post-hoc latent-change summaries.
The observed pattern is consistent with an early reframing of the domain clue, not with a deterministic ``algebraic'' mechanism.
}
\end{AIBox}
\caption{GPQA Diamond Q39: \ours reframes the clue as chromatographic separation.}
\label{fig:case_gpqa_q39}
\end{figure*}

\begin{figure*}[!t]
\begin{AIBox}[Case Study 5: IneqMath Q84: Correct Answer with a Flawed Proof]
\parbox[t]{\textwidth}{
\small

\PromptSection{Question}
Determine the correct relation for
\[
S = a\,\sqrt[3]{1{+}b{-}c} + b\,\sqrt[3]{1{+}c{-}a} + c\,\sqrt[3]{1{+}a{-}b}
\]
versus $1$, given $a+b+c=1$ with $a,b,c>0$.
Options include $\le$, $=$, $\ge$, etc.

\noindent\textcolor{AIgrayline}{\rule{\linewidth}{0.35pt}}

\PromptSection{Base Output}
Testing $a=b=c=\frac{1}{3}$ gives $S=1$.
The base model rewrites the terms using the constraint:
\[
1+b-c = a+2b, \qquad
1+c-a = b+2c, \qquad
1+a-b = 2a+c .
\]
These identities are valid under $a+b+c=1$.
The model then attempts AM-GM and ultimately concludes:
``Given these inequalities, the only consistent relationship is \textcolor{BaseErrRed}{$S=1$}.''
No valid option letter is extracted.

\noindent\textcolor{AIgrayline}{\rule{\linewidth}{0.35pt}}

\PromptSection{\ours Steered Output}
The \ours-steered reasoning trace uses auxiliary variables:
``Let $x=1+b-c$, $y=1+c-a$, $z=1+a-b$.
Note that $x+y+z=3$.
By the AM-GM inequality:
$\textcolor{BaseErrRed}{\sqrt[3]{1+b-c} \le \frac{x+y+z}{3} = 1}$.
Therefore $a\,\sqrt[3]{1+b-c} \le a$, and similarly for the other terms.
Summing: \textcolor{SteerFixGreen}{$S \le a+b+c = 1$}.''
Combined with $S=1$ at the symmetric point, the answer is \textcolor{SteerFixGreen}{A: $\le$}.

\noindent\textcolor{AIgrayline}{\rule{\linewidth}{0.35pt}}

\PromptSection{Key Difference}
\textcolor{BaseErrRed}{Base} concludes $S=1$, which holds at the symmetric point but not for all admissible inputs.
\textcolor{SteerFixGreen}{\ours} returns the correct relation $S\le1$, but its displayed proof is invalid: $x+y+z=3$ does not imply $x^{1/3}\le1$.
For example, $(a,b,c)=(0.1,0.8,0.1)$ gives $x=1.7>1$.

\PromptSection{Mathematical Check (Author Analysis)}
A valid proof uses the concavity of the cube root. With $x,y,z$ as defined above, $ax+by+cz=a+b+c=1$, so weighted Jensen gives
\[
S=a x^{1/3}+b y^{1/3}+c z^{1/3}\le (ax+by+cz)^{1/3}=1.
\]
Equality holds at $a=b=c=1/3$.

\PromptSection{Steering Info}
559 steering events and the first recorded steering events occur at generation steps 4, 5, and 6.
Top corrected dimensions: $z_4$ $\uparrow$ (variable extraction), $z_2$ $\uparrow$ (algebraic execution), $z_9$ $\downarrow$ (complexity evaluation).

\PromptSection{Reward-Signal Diagnostic}
The recorded gate activations locate latent interventions during generation. In this case, the steered output improves the final-answer relation while retaining an invalid inequality step. The case therefore illustrates a gap between final-answer correctness and proof correctness.

\PromptSection{Latent-Change Interpretation}
The largest average changes up-regulate $z_4$ and $z_2$, associated with variable extraction and algebraic execution, while down-regulating $z_9$, associated with complexity evaluation.
These changes accompany the use of auxiliary variables and the change in the final-answer relation, despite the remaining proof error.
These dimension names are post-hoc interpretability priors, not deterministic causal labels.
}
\end{AIBox}
\caption{IneqMath Q84: \ours returns the correct relation $S\le1$, but the displayed AM-GM argument is invalid. The author analysis provides a separate valid proof.}
\label{fig:case_ineqmath_q84}
\end{figure*}

\paragraph{Cross-case latent dimension patterns.}
Across the five cases, \ours reward-guided steering up-regulates $z_2$, accompanied by case-specific increases in $z_3$ or $z_4$ and decreases in $z_0$ or $z_9$.
The dimension $z_2$ appears as a top-corrected dimension in all five cases, indicating frequent updates to a dimension associated with algebraic step execution. The Q84 case shows that such updates can accompany a correct final answer while leaving a proof error unresolved.
This pattern holds across math competition tasks, graduate-level science questions, and inequality-proof settings.

\end{document}